\documentclass[lettersize,journal]{IEEEtran}
\usepackage{generic}
\usepackage{amsmath,amsfonts}
\usepackage{algorithmic}
\usepackage{algorithm}
\usepackage{array}
\usepackage[caption=false,font=normalsize,labelfont=sf,textfont=sf]{subfig}
\usepackage{textcomp}
\usepackage{stfloats}
\usepackage{url}
\usepackage{verbatim}
\usepackage{graphicx}
\usepackage{cite}
\usepackage{makecell}
\usepackage[section]{placeins}
\usepackage{flafter}
\usepackage[breaklinks=true,bookmarks=false,backref=page]{hyperref}
\usepackage{multirow}
\usepackage{threeparttable}
\usepackage{booktabs}
\usepackage{bbm}
\usepackage{todonotes}
\usepackage[export]{adjustbox}

\usepackage{pifont}
\newcommand{\xmark}{\ding{55}}%
\BeforeBeginEnvironment{appendices}{\clearpage}

\begin{document}


\title{MSPM: A Multi-Site Physiological Monitoring Dataset for Remote Pulse, Respiration, and Blood Pressure Estimation}

\author{
Jeremy Speth~\IEEEmembership{Student Member,~IEEE}, Nathan Vance~\IEEEmembership{Student Member,~IEEE}, Benjamin Sporrer, Lu Niu~\IEEEmembership{Student Member,~IEEE}, Patrick Flynn,~\IEEEmembership{Fellow,~IEEE}, and Adam Czajka,~\IEEEmembership{Member,~IEEE}
}



\maketitle




\begin{abstract}
Visible-light cameras can capture subtle physiological biomarkers without physical contact with the subject.
We present the Multi-Site Physiological Monitoring (MSPM) dataset, which is the first dataset collected to support the study of simultaneous camera-based vital signs estimation from multiple locations on the body. MSPM enables research on remote photoplethysmography (rPPG), respiration rate, and pulse transit time (PTT); it contains ground-truth measurements of pulse oximetry (at multiple body locations) and blood pressure using contacting sensors. We provide thorough experiments demonstrating the suitability of MSPM to support research on rPPG, respiration rate, and PTT. Cross-dataset rPPG experiments reveal that MSPM is a challenging yet high quality dataset, with intra-dataset pulse rate mean absolute error (MAE) below 4 beats per minute (BPM), and cross-dataset pulse rate MAE below 2 BPM in certain cases. Respiration experiments find a MAE of 1.09 breaths per minute by extracting motion features from the chest. PTT experiments find that across the pairs of different body sites, there is high correlation between remote PTT and contact-measured PTT, which facilitates the possibility for future camera-based PTT research.
\end{abstract}
\begin{IEEEkeywords}
Camera-Based Vitals, Physiological Monitoring, Remote Photoplethysmography
\end{IEEEkeywords}
\section{Introduction}

\begin{figure}
    \centering
    \includegraphics[width=.7\linewidth]{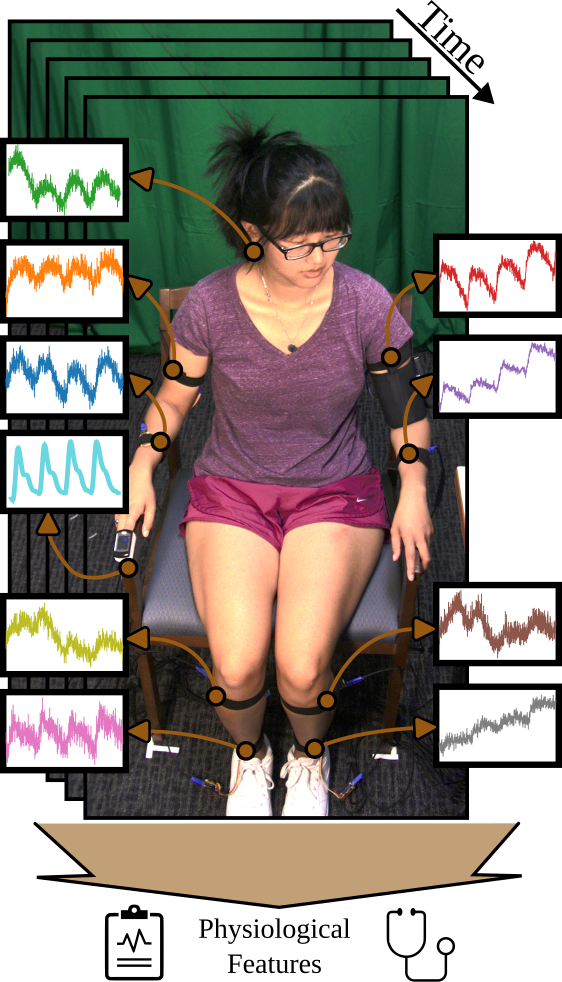}
    \caption{Key novelty of MSPM: Ground truth  PPG signal collected across multiple sites on the body with full-body video for its full utilization.}
    \label{fig:overview}
\end{figure}

\IEEEPARstart{C}{amera-based} vital signs estimation is a topic of growing interest. The vital signs of interest include heart rate measured via remote photoplethysmography (rPPG)~\cite{mcduff2015survey} or remote ballistocardiography (rBCG)~\cite{hassan2017video}, respiration rate~\cite{Zhan2020Resp}, blood pressure~\cite{gesche2012continuous,Shao_2017,Adachi2019,Iuchi_2022_CVPR}, and blood oxygenation~\cite{VanGastel2016,Verkruysse2017,VanGastel2018}. Remote vital signs estimation provides unobtrusive sensing and inexpensive deployment, enabling new applications where contact sensors would be impractical or impossible (such as monitoring automobile drivers for signs of drowsiness~\cite{xu2023ivrr}, screening airline passengers for infectious diseases~\cite{khanam2021noncontact}, monitoring prematurely born babies whose skin is too fragile for contact sensors~\cite{van2018near}, and using these vital signs in affective computing~\cite{yu2021facial}).

While many potential applications for remote vital signs estimation could employ any visible body part as an imaging site,  much of the rPPG literature contains only facial imaging. 
Furthermore, the majority of rPPG research has focused on robust heart rate estimation, which is only one physiological biomarker of interest.
In fact, we believe that the most interesting characteristic of camera-based measurement is not the noncontacting nature, but rather the ability to take many simultaneous measurements at different body locations from a single device.
As imaging hardware and algorithmic robustness improve, an opportunity will emerge for remote vital signs research to expand its scope to include the dynamics of blood flow throughout the body.

\IEEEpubidadjcol

In this paper we present the Multi-Site Physiological Monitoring (MSPM) dataset, consisting of RGB video taken at three different angles and covering the entire body, NIR video of the subjects' eyes, PPG data collected at 10 different sites across the body, cuff-based blood pressure readings, blood oxygenation (SpO2), and induced respiration. We further capture subjects performing various activities such as watching videos (including a video crafted to be an adversarial attack for rPPG~\cite{Speth_2022_WACV}), playing a computer game, undergoing a guided respiration exercise, and holding their breath to induce a change in blood pressure and oxygenation.

The main contributions of this paper are as follows:

\begin{itemize}
    \item The first rPPG dataset to our knowledge containing full-body video.
    \item The first rPPG dataset to our knowledge with PPG ground truth collected across multiple sites on the body.
    \item Canonical experiments demonstrating the utility of the MSPM dataset for rPPG, respiration, and PTT research.
\end{itemize}

A visualization of the novelty of the MSPM dataset is presented in Figure \ref{fig:overview}.

This work is a significant extension on the PTT research published in \cite{niu2023full}, in which a subset of MSPM was used for preliminary PTT experiments.
\section{Related Work}

\subsection{Camera-Based Physiological Measurement}
Simultaneous improvements to camera sensors and computer vision algorithms has allowed the field of camera-based physiological measurement to grow rapidly over the last decade~\cite{McDuff2023}.
Examples of some of the physiological measurements of interest include heart rate (which we describe in the next section), respiration rate~\cite{Zhan2020Resp}, blood pressure~\cite{Shao_2017,Adachi2019,Iuchi_2022_CVPR}, blood oxygenation~\cite{VanGastel2016,Verkruysse2017,VanGastel2018}, and pulse transit time~\cite{Shao_2014,Patil2019}.
Respiration is typically estimated by tracking motion of the body~\cite{Bartula2013} in regions such as the chest~\cite{Zhan2020Resp} or neck~\cite{Massaroni2018}.
Interestingly, pressure from respiration modulates the volume of blood in arterial sites, so it can also be estimated from the rPPG signal~\cite{Mirmohamadsadeghi2016}.

Blood oxygen saturation (SpO2) is another desirable vital sign to estimate remotely.
The optical mechanisms to estimate SpO2 require at least two wavelengths, ideally where the absorption coefficients of oxyhemoglobin and deoxyhemoglobin are separated.
There have been many successful attempts at camera-based SpO2 estimation with handcrafted and optically-informed approaches~\cite{VanGastel2016,Verkruysse2017,VanGastel2018}, but these experiments have been limited to very controlled scenarios.
One of the challenges in developing data-driven algorithms is the lack of variability in SpO2, which usually fluctuates between 95\% and 99\%.
In Sec.~\ref{sec:dataset}, we show that the MSPM dataset contains unique SpO2 variation due to a breath-holding activity.

Many researchers have begun pursuing camera-based blood pressure (BP) estimation~\cite{Natarajan2022}.
The physiological impacts of blood pressure changes can yield measurable optical changes.
Promising features indicative of blood pressure may be contained in the morphology of the rPPG waveform itself~\cite{Hosanee2020,Curran2023}.
However, the primary feature used in past studies of BP estimation is pulse transit time (PTT)~\cite{Shao_2017,Adachi2019,Iuchi_2022_CVPR}.

PTT is the measurement of the time lag of the pulse wave between two locations on the body, which is substantially easier to measure from a camera than from contact sensors~\cite{Shao_2014}.
The theoretical relationship between PTT and BP has been well-studied via arterial wave propagation models~\cite{Mukkamala2015}.
Assuming certain parameters such as arterial compliance, blood viscosity, and the arterial length are constant over short time periods, the PTT is inversely related to the BP.
Intuitively, a higher BP will result in faster propagation of blood through the arteries, causing a decrease in the PTT.
However, it is difficult to model the complex branching and tapering in the arterial tree that causes wave reflections, so PTT is typically weakly correlated with BP in experimental settings~\cite{Mukkamala2015,Gao2016,Block2020}.
Section~\ref{sec:multisite} describes experiments on PTT estimation using the MSPM dataset, and is a significant extension on \cite{niu2023full} by including statistical tests demonstrating the significance of our results.

\subsection{Remote Photoplethysmography (rPPG)}
rPPG is a technique for non-contact estimation of the blood volume pulse from digital videos of well-illuminated skin.
As the blood volume in microvasculature changes with each heart beat, the diffuse reflection of incident illumination changes due to the strong light absorption of hemoglobin. The observable changes are very subtle, even in modern camera sensors, with researchers suspecting that they are ``sub-pixel'' in amplitude~\cite{McDuff2023}. Initial approaches to infer the underlying PPG waveform from rPPG data utilized only the green color channel due the absorptive qualities of hemoglobin~\cite{Verkruysse2008}. Linear color transformation approaches~\cite{DeHaan2013,DeHaan2014,Wang2017,Wang2019} are still used as strong baselines due to their robustness and cross-dataset performance.

Some deep learning approaches regress the pulse rate directly, rather than a pulse or oxygenation waveform.~Niu \etal~\cite{Niu2018,Niu2020} passed spatial-temporal maps to ResNet-18 followed by a gated recurrent unit to predict the pulse rate. While the model is accurate on benchmark datasets, it lacks any measure of confidence or signal quality.

The most common deep learning approaches regress the pulse waveform values over a video~\cite{Chen2018,Yu2019,Liu_MTTS_2020,Lee_ECCV_2020,Lu2021,Speth_CVIU_2021,Zhao_2021,Yu_2022_CVPR}. Several approaches use frame differences to estimate the waveform derivative~\cite{Chen2018,Liu_MTTS_2020,Zhao_2021}. Many other approaches leverage spatiotemporal features, and can process video clips end-to-end~\cite{Yu2019,Lee_ECCV_2020,Lu2021,Speth_CVIU_2021,Yu_2022_CVPR}. There are numerous advantages to producing a full waveform, including the ability to extract unique cardiac features such as atrial fibrillation~\cite{Liu_JBHI_2022,Wu2023}, and the ability to use signal quality as a proxy for model confidence.

\subsection{Datasets for Remote Photoplethysmography}

Several datasets have been collected to study and evaluate rPPG techniques. MAHNOB-HCI~\cite{5975141} and PURE~\cite{Stricker2014} contain face videos with limited or controlled head movements. Estepp et al. collected AFRL, in which an array of cameras capture and mitigate head motion~\cite{estepp2014recovering}. Zhang et al. collected MMSE-HR~\cite{Zhang_2016_CVPR}, consisting of face videos after inducing emotional responses, as well as its extension intended for remote blood pressure research, BP4D+~\cite{zhang2016multimodal}. Heusch et al. collected COHFACE to promote reproducibility in rPPG research~\cite{Heusch2017}. Niu et al. collected VIPL-HR in both visible and near-infrared light~\cite{Niu2018}. Li et al. collected OBF, a dataset of both healthy subjects and subjects with Atrial Fibrillation~\cite{li2018obf}. UBFC-rPPG~\cite{Bobbia2019} contains subjects under stress from mathematical games, while UBFC-Phys~\cite{sabour2021ubfc} contains subjects undergoing social stress. DDPM~\cite{Speth_IJCB_2021, Vance2022} was the first long-form dataset containing unconstrained facial movements. Synthetic face rPPG datasets containing avatars have also been released~\cite{mcduff2022scamps,Kadambi2022}. Tang et al. released MMPD, a large dataset with 11 hours of video collected on mobile devices~\cite{tang2023mmpd}. As rPPG becomes more common for ubiquitous health monitoring, it is important to understand the limitations presented by partial or total occlusion of the face, as well as the corresponding opportunities when skin regions {\em not} on the face are visible. 

\begin{table}
    \centering
    \caption{Publicly available datasets for rPPG. *BP4D+ uses continuous blood pressure measurement rather than PPG.}
    \setlength{\tabcolsep}{4.5pt}
    \resizebox{0.99\columnwidth}{!}{%
    \begin{tabular}{lcccc}
        \toprule
        Dataset & \# Subjects & Minutes & \makecell{ \# PPG \\ Sites} & \makecell{ Full \\ body } \\
        \midrule
        MAHNOB-HCI~\cite{Soleymani2012} & 27 & 264 & 1 & \xmark \\
        PURE~\cite{Stricker2014} & 10 & 60 & 1 & \xmark \\
        UBFC-rPPG~\cite{Bobbia2019} & 43 & 70 & 1 & \xmark \\
        UBFC-Phys~\cite{sabour2021ubfc} & 56 & 504 & 1 & \xmark \\
        MMSE-HR~\cite{Zhang_2016_CVPR} & 40 & 102 & 1* & \xmark \\
        BP4D+~\cite{zhang2016multimodal} & 140 & 1285 & 1* & \xmark \\
        COHFACE~\cite{Heusch2017} & 40 & 160 & 1 & \xmark \\
        VIPL-HR~\cite{Niu2018} & 107 & 1150 & 1 & \xmark \\
        DDPM~\cite{Speth_IJCB_2021,Vance2022} & 93 & 776 & 1 & \xmark \\
        MMPD~\cite{tang2023mmpd} & 33 & 660 & 1 & \xmark \\
        \textbf{MSPM} & \textbf{103} & \textbf{1480} & \textbf{10} & \ding{52} \\

        \bottomrule
    \end{tabular}}
    \label{tab:rppgdatasets}
\end{table}

We examine and compare our Multi-Site Physiological Monitoring (MSPM) dataset to commonly used rPPG datasets, PURE, UBFC, and DDPM, and BP4D+. Details for these datasets are given in Table \ref{tab:rppgdatasets}.

\subsubsection{PURE}

The PURE dataset is a small rPPG dataset consisting of ten subjects. Of the five datasets examined, it has the lowest average heart rate and the lowest within-subject heart rate standard deviation, meaning that subjects maintained a relatively consistent heart rate.

\subsubsection{UBFC-rPPG}

The UBFC-rPPG dataset contains 43 subjects playing a time-sensitive mathematical game. Due to the heightened physiological response seen in subjects from this scenario, UBFC-rPPG has a high average heart rate and more heart rate variability than PURE.

\subsubsection{DDPM}

The DDPM dataset is an interview dataset comprised of 93 interviewee subjects in which the interviewee attempts to deceive the interviewer on selected questions. Each session is comprised of multiple questions and answers with recorded sessions lasting between 8 and 11 minutes. Since the collection protocol called for a mock interrogation with forced deception, it elicited stress in the subjects which in turn lead to a dataset with high heart rate variability.

\subsubsection{BP4D+}

The BP4D+ database is a Multimodal Spontaneous Emotion (MMSE) corpus that includes facial expressions, thermal, 2D and 3D dynamics, and physiological data. The physiological data, consisting of heart rate, blood pressure, respiration rate, and skin conductivity (EDA), was collected via a vital sign sensor. As a result, BP4D+ uses a continuous blood pressure measurement rather than PPG via a pulse oximeter, as is used in the other datasets. BP4D+ consists of 140 subjects who performed a total 10 tasks from 4 different approach categories: social interview, film watching, physical experience, and controlled activities. Each task was designed to trigger a different emotion.

\section{Dataset}
\label{sec:dataset}

\subsection{Metadata}

Data suitable for camera-based physiological measurement research was collected from 103 participants.  All data collection was performed under a human subjects research protocol approved by the University of Notre Dame's Human Subjects Institutional Review Board (protocol number 22-05-7249). A subset (the ``PTT subset'') of the data came from 87 subjects, for which all sensors and cameras worked without error. The pulse transit time (PTT) experiments described below were conducted with this subset of data.

Across the entire dataset the minimum age was 18 years old while the maximum was 58. The average age was 24.46 with a standard deviation of 8.02. For the PTT subset, the minimum and maximum ages remained the same while the average age increased to 24.64 with a standard deviation of 8.36.

The gender dispersion across the entire dataset was 58.3\% female and 39.8\% male. It was 57.5\% female and 40.2\% male for the PTT subset. The top three ethnicities for the entire dataset were White at 66\%, East Asian at 13.6\%, and Black at 4.9\%. For the PTT subset the top three enthnicities remained the same with 65.5\% White, 12.6\% East Asian, and 4.6\% Black.

The maximum BMI recorded across both the entire dataset and PTT subset was 41.53. The minimum for the entire dataset was 17.58 and for the PTT dataset was 17.78. The entire dataset had an average BMI of 23.79 with a standard deviation of 4.44. The PTT subset had an average BMI of 23.82 with a standard deviation of 4.34.

There was only one subject in the entire dataset who smoked. This subject was not a member of the PTT subset. For the entire dataset, 76.7\% of people had not consumed caffeine in the two hours prior to the data collection event. This percentage increased slightly to 77\% for the PTT subset. For the entire dataset, 23.3\% of the participants wore some type of makeup, while 24.14\% did for the PTT subset. Of those wearing makeup, mascara was the most frequently applied cosmetic type in both the entire dataset and the PTT subset with 83.33\% and 85.71\% of subjects (respectively) wearing mascara.

\subsection{Apparatus}

\begin{figure}
    \centering
    \includegraphics[height=0.85\linewidth,valign=t]{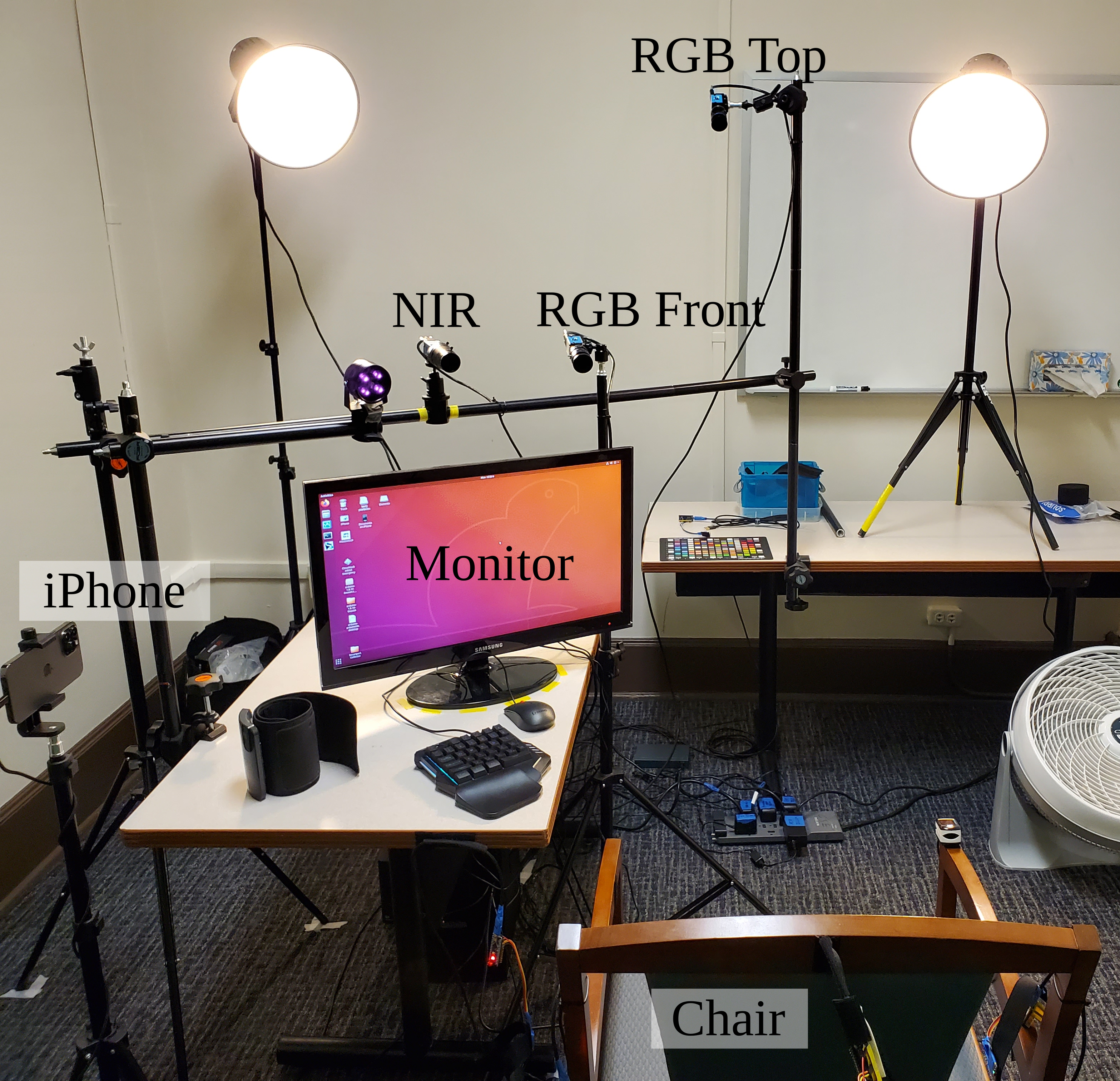}
    \caption{Placement of sensors in the collection room.}
    \label{fig:room}
\end{figure}

Subjects sat in a 45 cm tall chair located 64 cm from an LCD monitor, which is a realistic and recommended viewing distance~\cite{rempel2007effects}. The overall set of sensors consisted of:

\begin{enumerate}
\item a DFK 33UX290 {\bf RGB camera} (``RGB Top" in Figure \ref{fig:room}) from The Imaging Source (TIS) operating at 90 FPS with a resolution of 1920 $\times$ 1080 pixels, positioned 138 cm horizontally from the center of the subject's chair and 166 cm vertically from the ground; 
\item a DFK 33UX290 {\bf RGB camera} (``RGB Front" in Figure \ref{fig:room}) from TIS operating at 30 FPS with a resolution of 1920 $\times$ 1080 pixels, positioned 94 cm horizontally from the center of the subject's chair and 123 cm vertically from the ground; 
\item a DMK 33UX290 monochrome camera from TIS with a bandpass filter to capture {\bf near-infrared images} (730 to 1100 nm) at 30 FPS and 1920 $\times$ 1080 pixels, positioned 100 cm horizontally from the center of the subject's chair and 121 cm vertically from the ground; 
\item an iPhone 13 Pro Max {\bf mobile phone} that yielded 640 $\times$ 480 pixel images at 30 FPS, positioned 94 cm horizontally from the center of the subject's chair and 94 cm vertically from the ground; 
\item a FDA-certified Contec CMS50EA {\bf pulse oximeter} that provided a 60 samples/second SpO2 and heart rate profile; 
\item Nine Maxim Integrated MAX30101 contact-PPG {\bf pulse sensors} that recorded red and near-infrared signals at 400 samples/second;
\item a FDA-certified Omron Evolv cuff-based blood pressure monitor.
\end{enumerate}

\begin{figure}
    \centering
    \includegraphics[height=0.85\linewidth,valign=t]{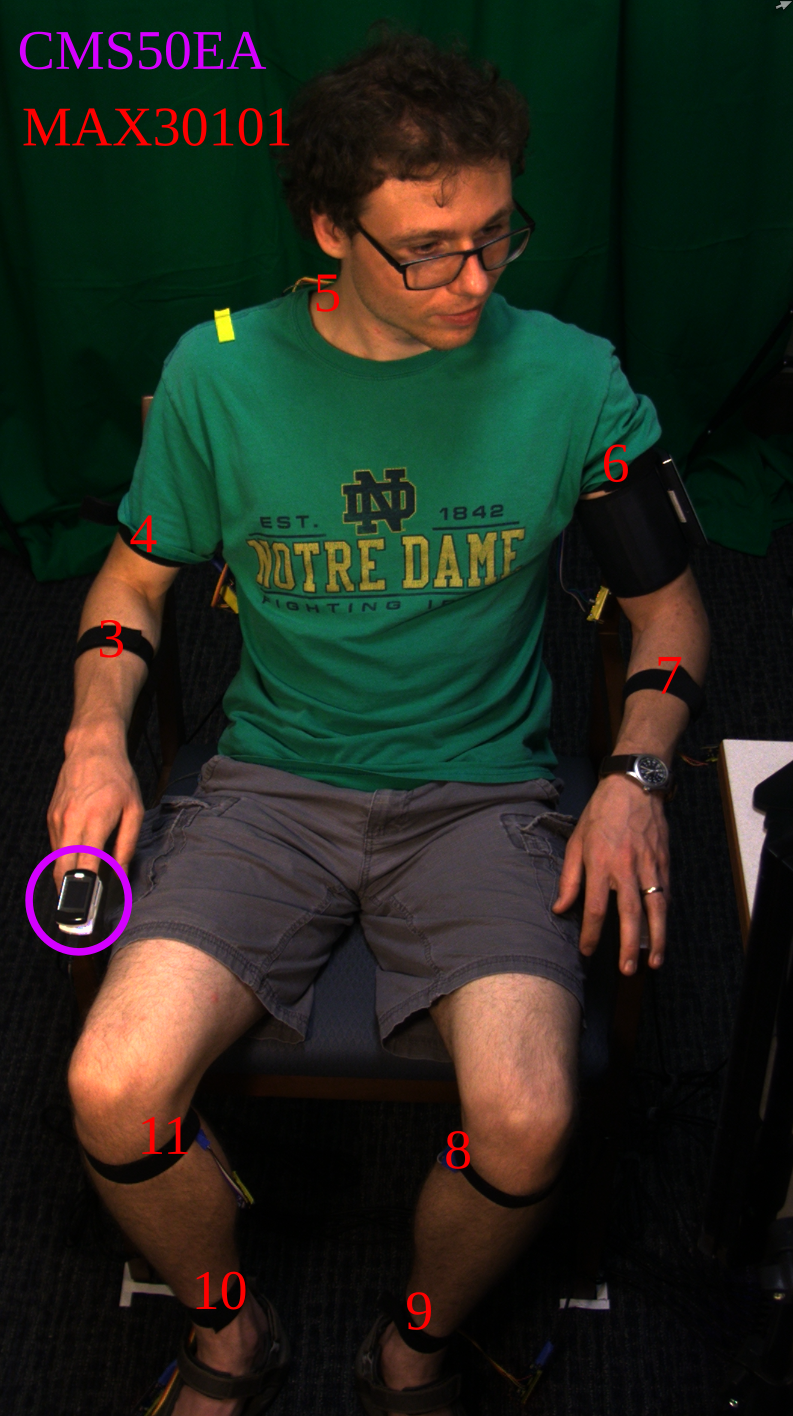}
    \caption{PPG sensors were attached to the subject's neck and limbs.}
    \label{fig:sensorplacement}
\end{figure}

\begin{figure*}
    \centering
    \captionsetup[subfloat]{labelfont=footnotesize,textfont=footnotesize}
    \subfloat[][RGB front]{\includegraphics[height=.16\textheight]{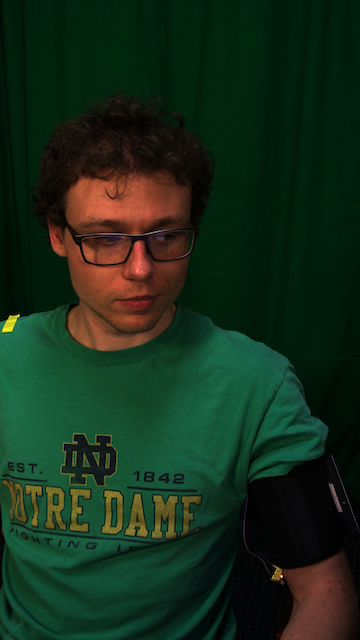}\label{fig:rgbfront}}
    \hfil
    \subfloat[][RGB top]{\includegraphics[height=.16\textheight]{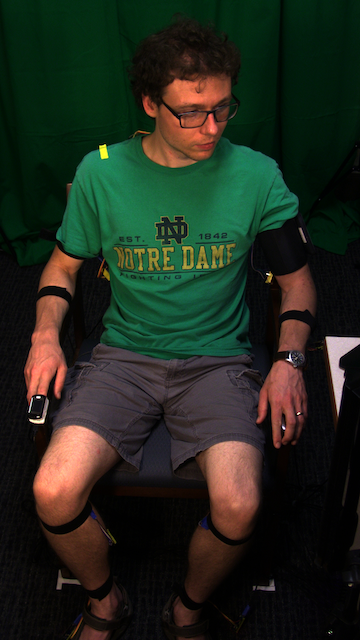}\label{fig:rgbtop}} 
    \hfil
    \subfloat[][iPhone]{\includegraphics[height=.16\textheight]{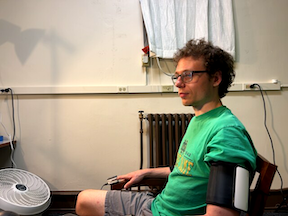}\label{fig:iphone}} 
    \hfil
    \subfloat[][NIR]{\includegraphics[height=.16\textheight]{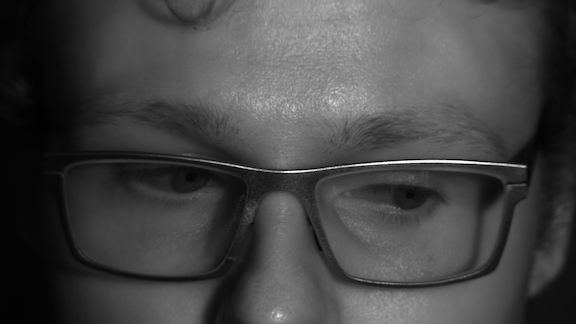}\label{fig:nir}}
    \caption{Imaging modalities in MSPM include RGB video from the front \ref{fig:rgbfront}, overhead \ref{fig:rgbtop}, and profile \ref{fig:iphone}, and NIR video of the eyes \ref{fig:nir}.}
    \label{fig:imagingmodalities}
\end{figure*}

The placement of sensors in the room is shown in Figure \ref{fig:room}. The Contec CMS50EA pulse oximeter and 9 (numbered 3 through 11 in correspondence with their labels in the database) MAX30101 contact-PPG pulse sensors were attached to subjects with a strap, as indicated in Figure \ref{fig:sensorplacement}. A sample from each imaging modality
is given in Figure \ref{fig:imagingmodalities}.


\subsection{Collection Protocol} \label{sec:collectionprotocol}


\begin{figure}
    \centering
    \includegraphics[width=\linewidth]{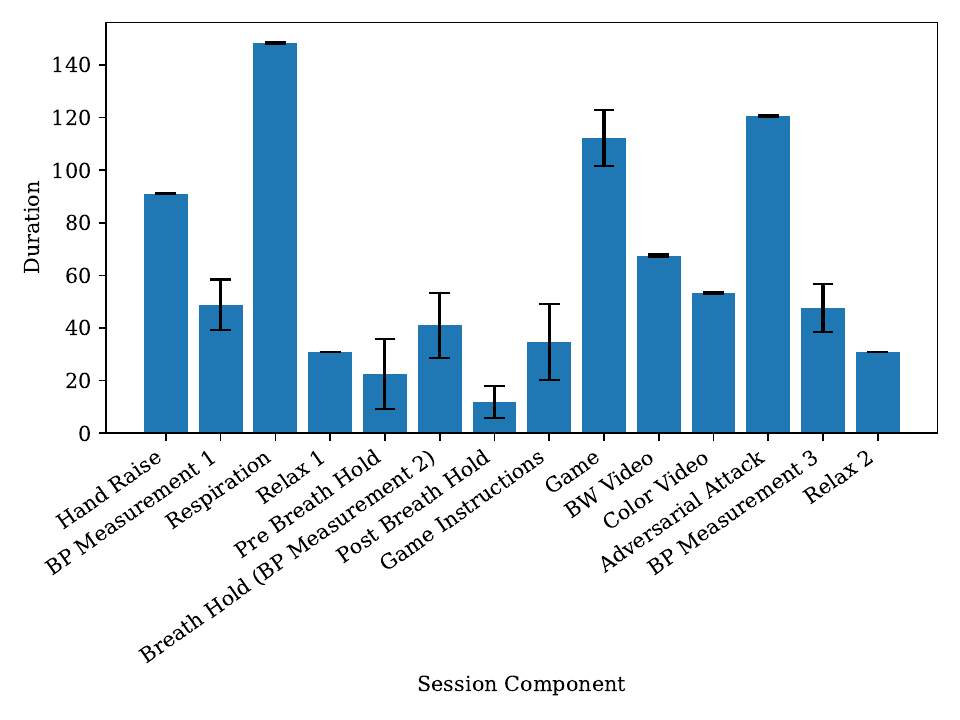}
    \caption{Duration of each activity, averaged across sessions. The Hand Raise, Relax segments, and videos (Respiration, BW Video, Color Video, Adversarial Attack) automatically advanced on completion thereby resulting in low duration variability.}
    \label{fig:timestamps}
\end{figure}

\begin{figure}
   \centering
   \includegraphics[width=\columnwidth]{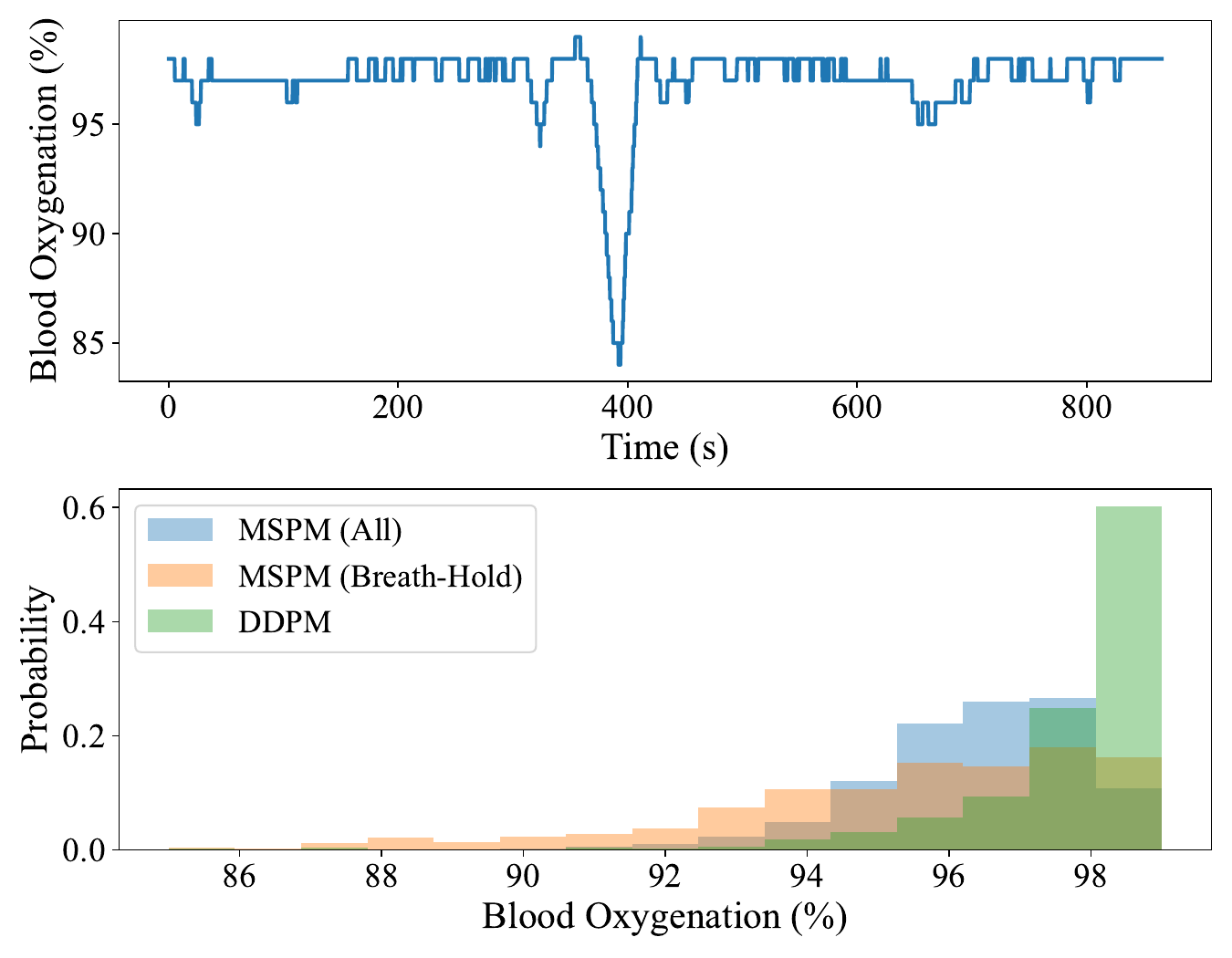}
   \caption{The top figure shows oxygenation for a single subject over the course of a session, where a drop occurs during the breath-holding activity. The bottom figure shows a comparison of blood oxygenation values between the MSPM and DDPM datasets. Data for the breath-holding activity is offset by 30 seconds to accommodate the physiological delay in decreased oxygenation.}
   \label{fig:spo2}
 \end{figure}

\begin{figure}
     \centering
     \includegraphics[width=0.8\columnwidth]{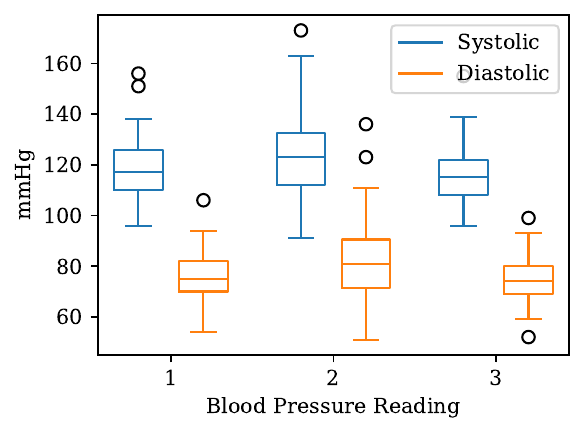}
     \caption{Systolic and diastolic blood pressure measurements were collected at three different points in each session as shown.}
     \label{fig:bp}
 \end{figure}

During the collection, the subjects completed the following sequence of prompted activities:

\begin{enumerate}
    \item \textbf{Hand Raise:} The subjects raised their left hands such that the palm faced the camera for 90 seconds.
    \item \textbf{BP Measurement:} The first blood pressure measurement was collected, taking 49 seconds on average.
    \item \textbf{Respiration} The subjects followed a guided respiration video. The first 28 seconds explained how the subject should follow a cue on the screen for inhaling and exhaling. The breathing exercise then occurred for 120 seconds with the respiration rate ranging from 10 to 20 breaths per minute.
    \item \textbf{Relax:} The subjects were instructed to relax for 30 seconds.
    \item \textbf{Pre Breath Hold:} The subjects were asked how long they would be comfortable attempting to hold their breath (the response was used to time the blood pressure measurement such that it was completed near the end of the breath hold). The researcher then counted down from 5 to begin the breath hold. On average this exchange lasted 22 seconds.
    \item \textbf{Breath Hold (BP Measurement):} The subjects held their breath and the second blood pressure measurement was simultaneously collected. Subjects elected to hold their breath for 41 seconds on average. An outcome of this activity is an induced SpO2 fluctuation as shown in Figure~\ref{fig:spo2}, resulting in increased SpO2 variability compared with other rPPG datasets. Additionally, it induced an elevated blood pressure for this second blood pressure reading, as shown in Figure~\ref{fig:bp}.
    \item \textbf{Post Breath Hold:} The researcher recorded the results of the breath hold while the subjects recovered, taking an average of 12 seconds.
    \item \textbf{Game Instructions:} The subjects were instructed on the controls for a racing video game for an average of 35 seconds.
    \item \textbf{Game:} The subjects played a short racing game (the tutorial track from SuperTuxKart\footnote{https://supertuxkart.net}). The game duration was capped at 2 minutes, but subjects required 112 seconds on average due to some subjects completing the course before time ran out.
    \item \textbf{BW Video:} The subjects watched a 68 second long scene from \textit{It's A Wonderful Life}.
    \item \textbf{Color Video:} The subjects watched a 53 second long scene from \textit{Star Wars VI: Return of the Jedi}.
    \item \textbf{Adversarial Attack:} The subjects watched the screen pulsate between green and red light at 120 pulsations per minute and at an increasing intensity throughout an attack lasting 120 seconds. We used the same color pattern for the attack as was used in~\cite{Speth_2022_WACV}.
    \item \textbf{BP Measurement:} The third blood pressure measurement was collected, taking 48 seconds on average.
    \item \textbf{Relax:} The subjects were instructed to relax for 30 seconds.
\end{enumerate}

The timing of these activities is summarized in Figure \ref{fig:timestamps}.

 \subsection{Signal Curation}

In order to evaluate current state of the art rPPG approaches, a single global pulse rate was required. We combined the pulse rates from the multiple contact sensors to produce an estimate for this global pulse rate. Due to varying and unpredictable body movement throughout a given session, varying levels of noise are present in the contact PPG signals recorded from the nine sites with the MAX30101 sensors. Since movement may have been isolated to particular regions of the body, the true pulse signal is most likely present in at least one signal at each instantaneous time.

\begin{figure}[ht]
    \centering
    \includegraphics[width=\linewidth]{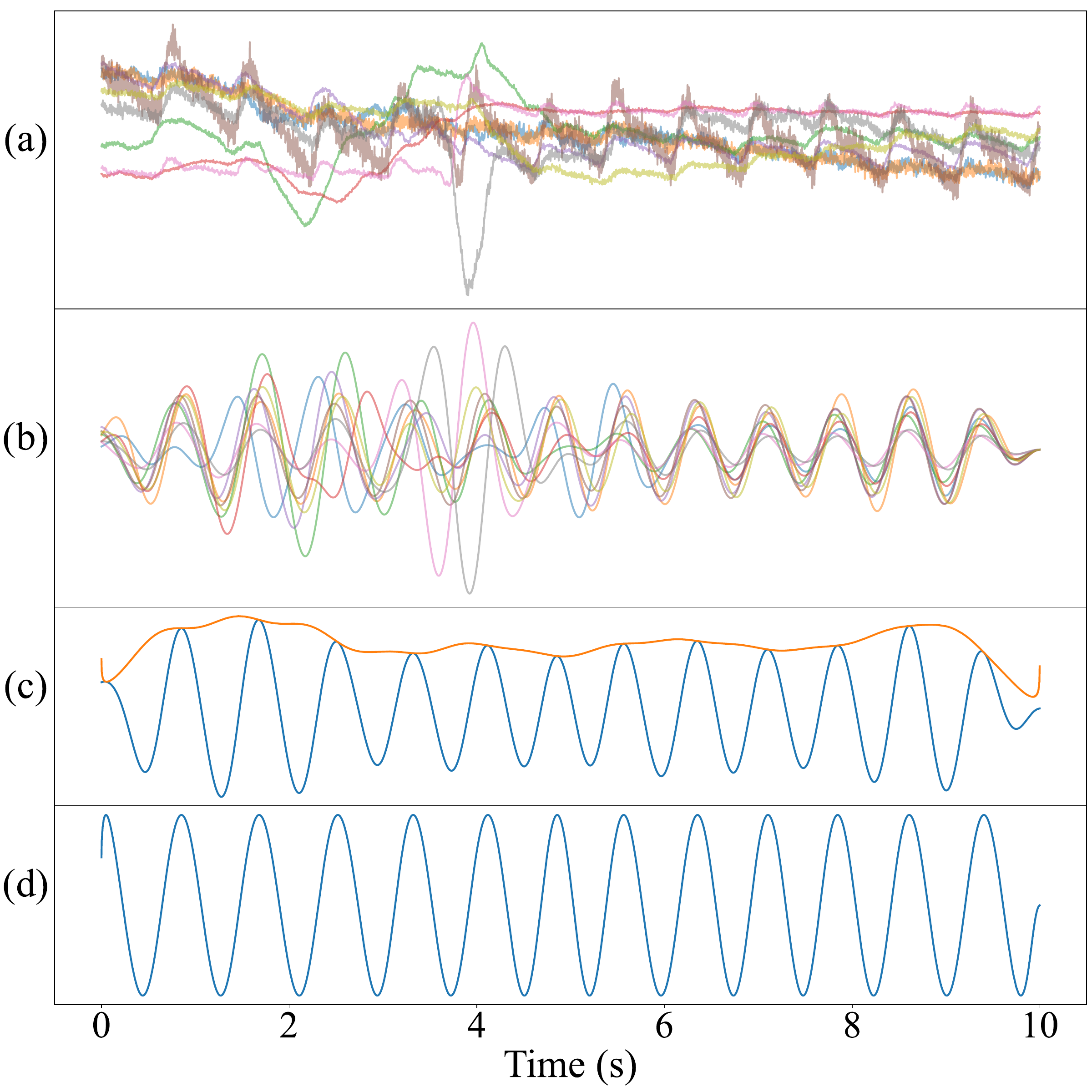}
    \caption{Processing of the contact PPG waveforms to produce a robust pulse rate. (a) The z-normalized waveforms from all sensors. (b) Bandpass filtered signals around the fingertip oximeter's pulse rate $\pm$ 30 bpm. (c) The signals were added together and the envelope is calculated. (d) The combined signal was divided by the envelope.}
    \label{fig:ground_truth}
\end{figure}

Taking this as an axiom, we combined the multiple pulse signals using a windowed bandpass filtering technique. To define the bounds of the narrow bandpass filter for the MAX30101 signals, we utilized the FDA-certified CMS50EA oximeter's pulse rate estimates. We designated the estimated pulse rate from the fingertip oximeter, $Y$, as our stable variable for each point in time. We then specified a padding around this value, $\Delta Y = 30$ bpm, and filtered the MAX30101 signals with a 2nd order Butterworth filter with lower and upper cutoffs of $Y - \Delta Y$ and $Y + \Delta Y$, respectively.

Specifically, for a sliding 10-second window with a stride of a single sample at 400 Hz, the waveforms underwent z-normalization, followed by filtering around the fingertip oximeter's pulse rate, then they were summed together into the combined waveform for that window. Finally, we calculated the envelope of the waveforms via the Hilbert transform and then divided the complete combined waveform by it. The complete process for combining signals is displayed in Fig.~\ref{fig:ground_truth}. Subplots (a) and (b) exhibit signal noise between seconds 2 and 5. Multiple sensors show evidence of this noise, but the combined global signal seen in subplot (d) is clean upon performing the techniques described.
\section{Experiments}

\subsection{Evaluation Metrics}
\label{sec:evaluation-metrics}
Pulse detection performance is analyzed by calculating the error between predicted and ground truth heart rates. The heart rate is calculated by applying a 10 second wide Short Term Fourier Transform (STFT), from which the maximum spectral peak between 0.66 Hz and 3 Hz (40 bpm to 180 bpm) is selected as the heart rate. Since the frequency domain suffers quantization effects, we pad the spectrum with zeros such that the quantization is reduced to 0.001 Hz.

\subsubsection{Mean Error (ME)}

The ME captures the bias of the method in BPM, and is defined as follows:

\begin{equation}
    ME = \frac{1}{N}\sum\limits_{i=1}^{N} (HR'_i - HR_i)
\end{equation}

Where $HR$ and $HR'$ are the ground truth and predicted heart rates, respectively. Each contained index is the heart rate obtained from the STFT window. $N$ is the number of STFT windows present.


\subsubsection{Mean Absolute Error (MAE)}

The MAE captures an aspect of the precision of the method in BPM, and is defined as follows:

\begin{equation}
    MAE = \frac{1}{N}\sum\limits_{i=1}^{N} |HR'_i - HR_i|
\end{equation}

\subsubsection{Root Mean Squared Error (RMSE)}

The RMSE is similar to MAE, but penalizes outlier heart rates more strongly:

\begin{equation}
    RMSE = \sqrt{\frac{1}{N}\sum\limits_{i=1}^{N} (HR'_i - HR_i)^2}
\end{equation}

\subsubsection{Waveform Correlation ($r_{wave}$)}

The waveform correlation, $r_{wave}$, is the Pearson's $r$ correlation coefficient between the ground truth and predicted waves.

\subsubsection{MXCorr}

The Maximum Cross Correlation (MXCorr) is calculated as the maximum Pearson's $r$ correlation for all correlation lags up to $\pm$1 second.

\subsection{Respiration}

\begin{figure}
   \centering
   \includegraphics[width=\columnwidth]{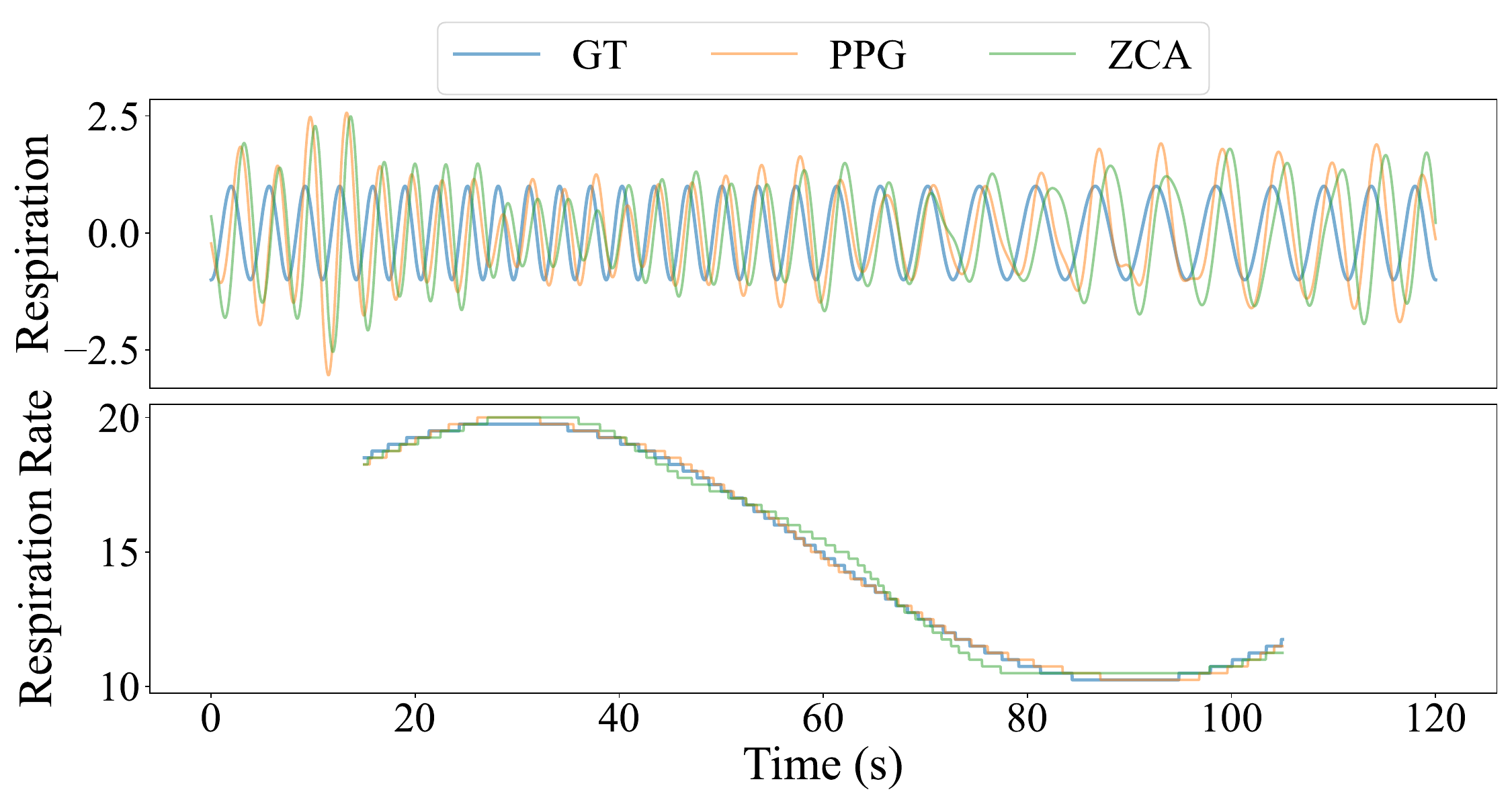}
   \caption{Comparison between remote respiration estimation and the ground truth for a single subject using ZCA on the vertical optical flow of the chest region~\cite{Zhan2020Resp}.}
   \label{fig:respiration}
 \end{figure}

Several useful health diagnostics can be inferred from a subject's breath. During our data collection events, inspired by the technique described by Zhan \etal~\cite{Zhan2020Resp}, we displayed a sinusoidal breathing pattern on the computer monitor which subjects followed with their breath, thus providing a ground truth for our respiration experiments. The breathing frequencies were modulated within 0.167-0.333 Hz (10-20 breaths per minute), which is considered a healthy range for adults~\cite{russo2017physiological}. Figure~\ref{fig:respiration} shows the ground truth respiration waveform and frequencies alongside the camera-based respiration estimated from a video.

\subsubsection{PPG-Based Respiration}
Multiple simultaneous PPG recordings contain many features for health monitoring, including respiration~\cite{Dehkordi2018}. We designed a simple approach to extract the respiration signal from the contact signals as a baseline for future studies exploring this topic. For each individual PPG signal, we normalized the signal to zero mean and unit variance for the length of the entire recording. Then, we applied a 3rd order Butterworth filter with lower and upper cutoffs of 6 and 24 breaths per minute, respectively. Finally, all of the bandpass-filtered signals were summed together for a final respiration estimate. Figure~\ref{fig:respiration} shows an example of the PPG-based prediction compared with the ground truth.

\subsubsection{Camera-Based Respiration}
As a baseline evaluation for remote respiration, we applied the best-performing approach from ~\cite{Zhan2020Resp} on the RGB Front videos from the dataset.
We first cropped each video by selecting the minimum and maximum $(x,y)$ locations of the ``upper-clothes" mask from the human parsing package~\cite{li2020self}.
A static bounding box was selected as the mean $(x,y)$ coordinates for the entire video.
The bounding box must be constant to avoid introducing motion during cropping.
Next, we applied dense optical flow within the bounding box using \cite{FarnebackOF}.
The vertical ($y$-axis) motion was calculated, and the horizontal ($x$-axis) movement was discarded.
The dense optical flow videos were then downscaled to $10 \times 10$ pixels, flattened to an $N \times 100$ matrix, then we performed zero-phase component analysis (ZCA).
From all of the projected components, we averaged the three with the highest signal-to-noise ratio (SNR).
The SNR was calculated as the sum of signal power between 10 and 20 breaths per minute, divided by the power outside those bounds.

Table~\ref{tab:respiration} shows the results over the whole MSPM dataset, where ME, MAE, and RMSE are presented in units of breaths per minute. Interstingly, we find that both the camera-based and contact approaches give very similar performance. Overall, both methods give practically useful results, with strong correlations near 0.8.

\begin{table}
    \centering
    \caption{Respiration results using both the contact-PPG sensors and video.}
    \setlength{\tabcolsep}{5.0pt}
    \resizebox{0.99\columnwidth}{!}{%
    \begin{tabular}{cccccc}
        \toprule
        Method & Sensor & ME & MAE & RMSE & $r$ \\
        \midrule
        PPG Filtering & Contact-PPG & 0.21 & 1.15 & 2.52 & 0.77 \\
        ZCA~\cite{Zhan2020Resp} & Lower RGB & -0.27 & 1.09 & 2.40 & 0.79 \\
        \bottomrule
    \end{tabular}}
    \label{tab:respiration}
\end{table}



\subsection{Multi-site Remote Photoplethysmography}
\label{sec:multisite}

Prior applications of rPPG focused mainly on the face because past video datasets only captured the facial region. Exploiting the MSPM dataset's expanded body coverage, we explored rPPG performance on non-face body regions such as the arms, legs, and hands. This section expands upon the results presented in ~\cite{niu2023full} by using all 103 subjects and all session components excluding the adversarial attack. We applied several rPPG approaches, including a chrominance-based technique (CHROM)~\cite{DeHaan2013}, plane-orthogonal-to-skin (POS)~\cite{Wang2017}, and RemotePulseNet (RPNet)~\cite{Speth_CVIU_2021} to estimate the heart rate from non-face regions.

\subsubsection{Region of Interest (ROI) Selection}
\begin{figure}
    \centering
    \captionsetup[subfloat]{labelfont=footnotesize,textfont=footnotesize}
    \subfloat[]{\includegraphics[width=0.45\linewidth,valign=t]{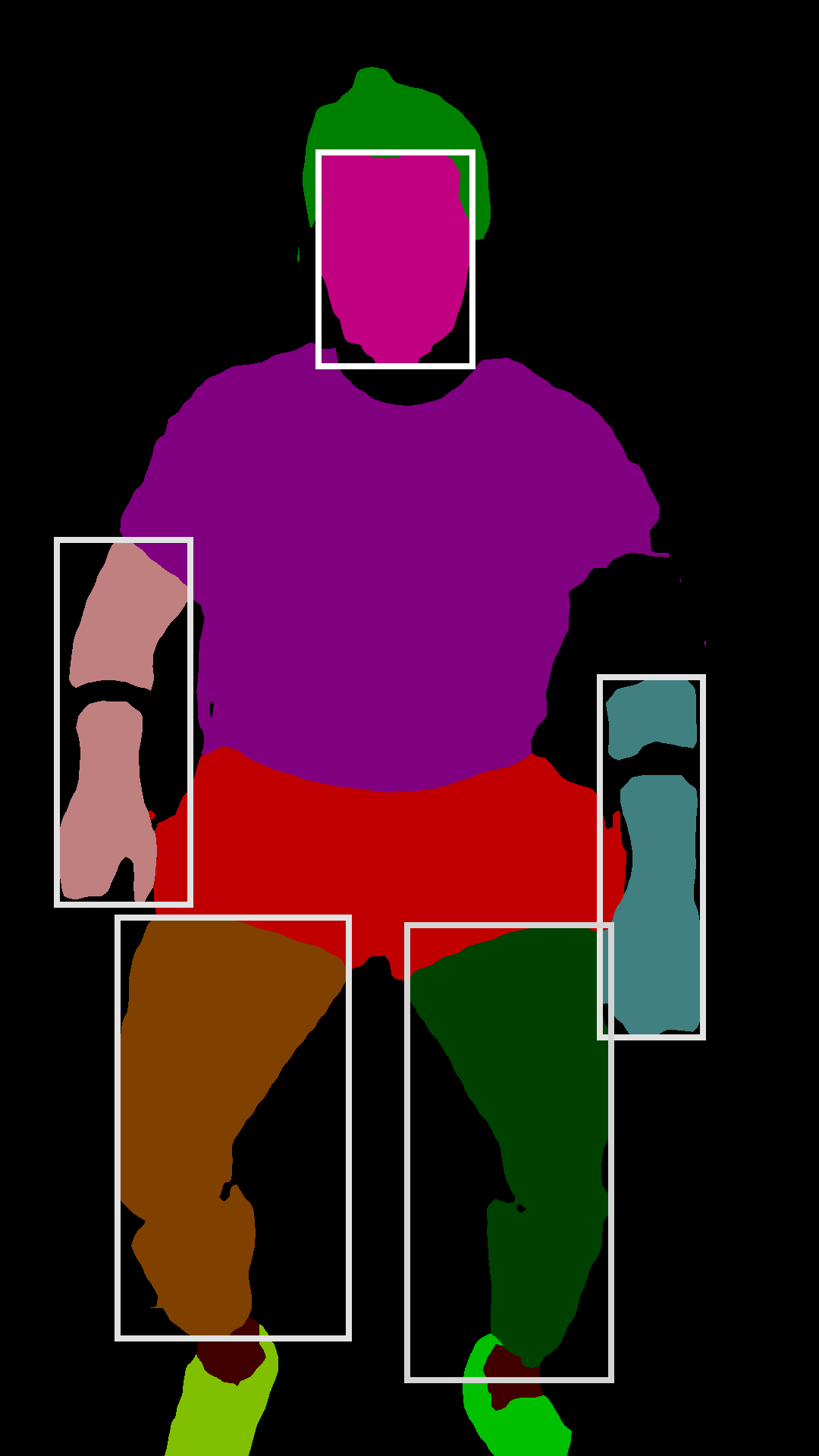}\label{subfig:schp_l}}
    \hfil
    \subfloat[]{\includegraphics[width=0.45\linewidth,valign=t]{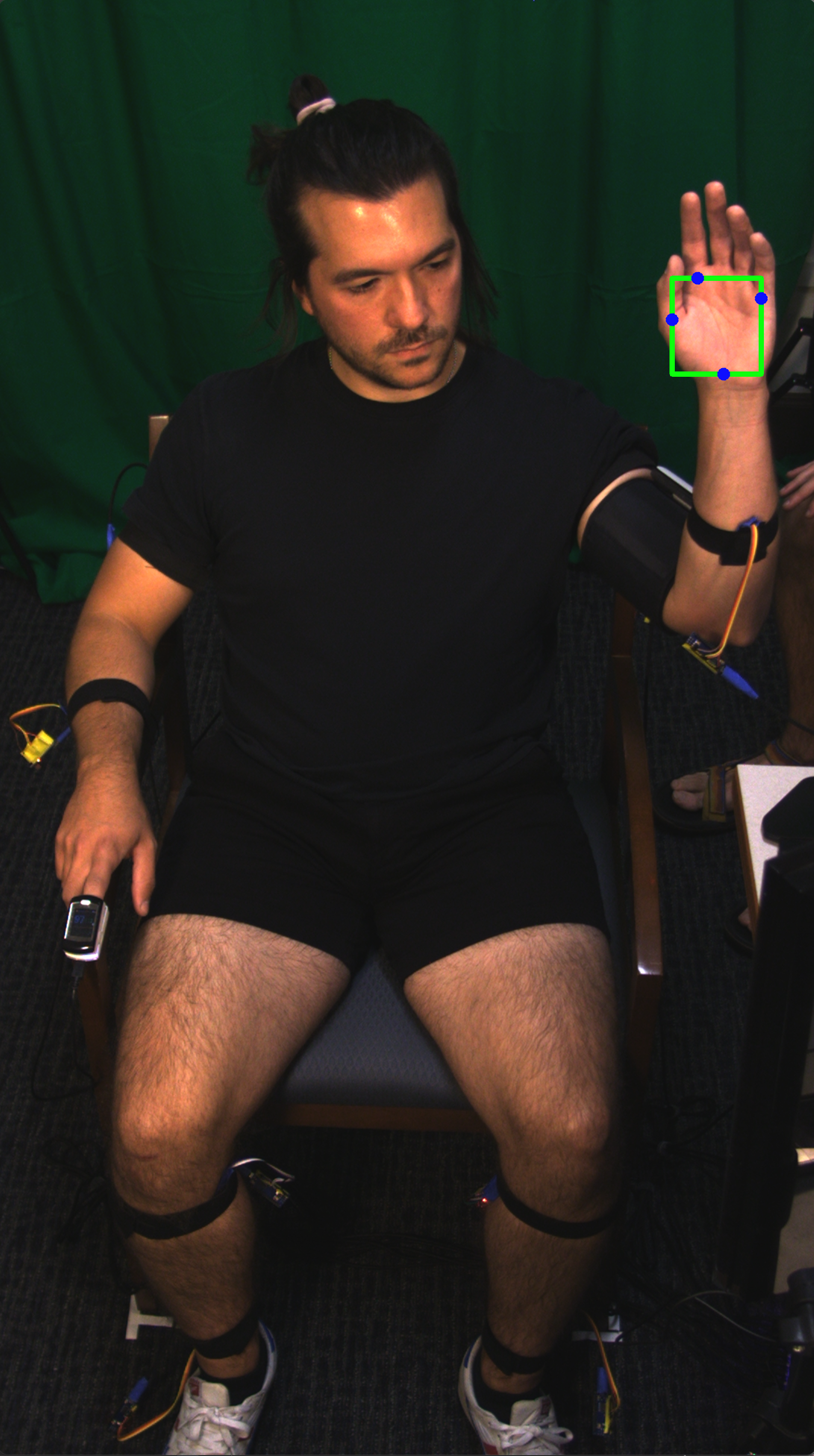}\label{subfig:bbox}}
    \hfil    
    \caption{\ref{subfig:schp_l} shows human body segmentation using SCHP~\cite{li2020self} and the associated bounding boxes as done in ~\cite{niu2023full}. \ref{subfig:bbox} shows MediaPipe Hand detection~\cite{Zhang2020MPHands} applied to detect 4 points (shown in blue) of the palm which are used to form a bounding box (shown in green). }
    \label{fig:schp-bbox}
\end{figure}

We selected skin pixels from the face, arms, and legs as the five primary ROIs.
Figure~\ref{fig:schp-bbox} shows the selected ROIs for full-body rPPG with the different methods~\cite{niu2023full}. We first applied Self-Correction Human Parsing (SCHP)~\cite{li2020self} to mask skin pixels from each arm, leg, and the face. For CHROM and POS, we spatially averaged the skin pixels in each mask. For RPNet, which takes fixed video inputs, we generated bounding boxes of each body part using minimum and maximum values of ($x, y$) locations of each mask, as shown by the white boxes in Fig.~\ref{fig:schp-bbox}. Similarly, for the ``hand raise" activity we created bounding boxes for the palms using four key points of the palm generated by MediaPipe Hand detection~\cite{Zhang2020MPHands}. Figure~\ref{subfig:bbox} shows an example palm bounding box used for CHROM, POS, and RPNet.

\setlength\tabcolsep{3.9pt}
\begin{table*}[htb!]
\caption{Pulse rate estimation results over different body sites. MAE: Mean Absolute Error; $r$: Pearson correlation coefficient.}
\fontsize{9.0}{9}\selectfont
\centering
\begin{tabular}{l cc cc cc cc cc | cccc}
\toprule
& \multicolumn{10}{c|}{All of session except adversarial attack video} & \multicolumn{4}{c}{Hand raise}\\
\cmidrule(lr){2-11}\cmidrule(lr){12-15}
\multirow{3}{*}{\begin{tabular}[c]{@{}l@{}}\\Methods\end{tabular}} &
\multicolumn{2}{c}{Face} &
\multicolumn{2}{c}{Right leg} &
\multicolumn{2}{c}{Left leg} &
\multicolumn{2}{c}{Right arm} &
\multicolumn{2}{c|}{Left arm} &
\multicolumn{2}{c}{Left arm} &
\multicolumn{2}{c}{Palm}\\ 
\cmidrule(lr){2-3}
\cmidrule(lr){4-5}
\cmidrule(lr){6-7}
\cmidrule(lr){8-9}
\cmidrule(lr){10-11}
\cmidrule(lr){12-13}
\cmidrule(lr){14-15}
& \begin{tabular}[c]{@{}c@{}}MAE\\ (bpm)\end{tabular} & $r$
& \begin{tabular}[c]{@{}c@{}}MAE\\ (bpm)\end{tabular} & $r$
& \begin{tabular}[c]{@{}c@{}}MAE\\ (bpm)\end{tabular} & $r$
& \begin{tabular}[c]{@{}c@{}}MAE\\ (bpm)\end{tabular} & $r$
& \begin{tabular}[c]{@{}c@{}}MAE\\ (bpm)\end{tabular} & $r$
& \begin{tabular}[c]{@{}c@{}}MAE\\ (bpm)\end{tabular} & $r$
& \begin{tabular}[c]{@{}c@{}}MAE\\ (bpm)\end{tabular} & $r$\\
\midrule
CHROM \cite{DeHaan2013} &
    3.66 & 0.72 &
    10.54 & 0.42 &
    11.12 & 0.39 &
    10.43 & 0.40 &
    12.74 & 0.29 &
    4.69 & 0.66 &
    5.55 & 0.64 \\
POS \cite{Wang2017} &
    \bf 2.03 & \bf 0.83 &
    \bf 6.91 & \bf 0.53 &
    \bf 7.39 & \bf 0.51 &
    \bf 4.48 & \bf 0.67 &
    \bf 8.96 & \bf 0.46 &
    \bf 3.72 & \bf 0.72 &
    \bf 4.31 & \bf 0.73 \\
RPNet \cite{Speth_CVIU_2021}\tnote{\textdagger} &
    3.49 & 0.74 &
    30.35 & 0.12 &
    31.52 & 0.09 &
    25.51 & 0.14 &
    26.26 & 0.11 &
    11.16 & 0.37 &
    7.21 & 0.49 \\
    
\bottomrule
\end{tabular}
\label{tab:within_dataset}
\end{table*}

\subsubsection{rPPG Signals Extraction}
For CHROM~\cite{DeHaan2013} and POS~\cite{Wang2017}, we spatially averaged skin pixels for each body part, resulting in 1D signals for the RGB channels which CHROM and POS then linearly combine to remove noise from subject motion and lighting. We also obtained deep learning rPPG results --- using the coordinates of bounding boxes of each body part, we applied bicubic interpolation to downsize each ROI to 64x64 pixels, which we fed to a pretrained RPNet model for rPPG predictions. We utilized models that were pretrained on DDPM~\cite{Vance2022} as it and the full-body video are both 90 fps.

\begin{figure}
    \centering
    \includegraphics[width=\linewidth]{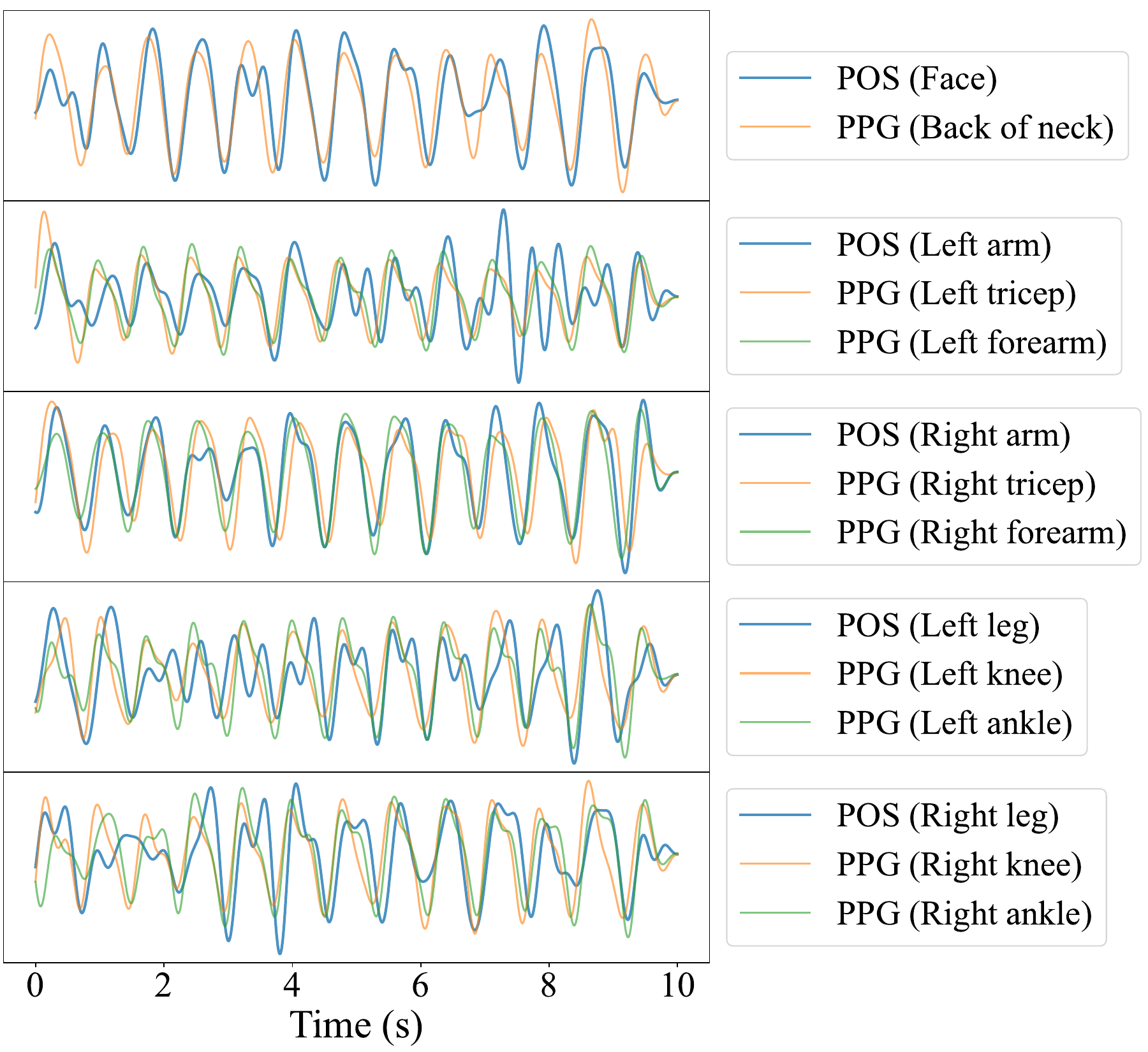}
    \caption{Pulse waveform predictions from POS overlayed with the nearest contact-PPG waveforms for a 10 second window.}
    \label{fig:result_waveforms}
\end{figure}

Remote pulse estimation over non-face regions is challenging due to it being non-glabrous skin, which results in a lower signal-to-noise ratio than the face. To improve the signal quality for all approaches, we applied a 4th order Butterworth bandpass filter with cutoff frequencies of 40 bpm and 180 bpm. Bandpass filtering was not used in the original POS and RPNet implementations, but we found that the POS estimates in particular contained high frequency noise before filtering.
 
\subsubsection{Multi-Site Pulse Rate Evaluation}


We calculated pulse rate and evaluated the performances of the three methods according to \ref{sec:evaluation-metrics}.
Table~\ref{tab:within_dataset} shows the results of rPPG experiments across different body parts, excluding the adversarial attack.
We additionally report results for the left arm and palm during the ``hand raise" activity since this is the only segment of each session where the palm of the hand is visible.
All of the rPPG estimators achieve the best performance in the face region.
We found that the RPNet model trained on the face did not transfer well to non-facial regions, indicating that the model learned to respond to spatial features of the face.
Additionally, the deep learning model may be overfit to the skin thickness, melanin concentration, and microcirculation present in the glabrous skin of the face.
The recuperated performance for RPNet on the palm (also with glabrous skin) helps justify this explanation.

The POS algorithm gives the best performance for all body regions in terms of both MAE and $r$. This is especially impressive given that it is a simple linear method.
It also shows that color changes from blood volume are similar over different skin thicknesses and underlying microvasculature. 
For POS and CHROM, the order of performance from best to worst is face, arms, then legs.
The palm also gives good performance for all approaches, due to physical similarities to the skin on the face.
Interestingly, as shown in \cite{niu2023full}, the errors generally occur as spikes of short duration rather than sustained periods of large offset. It is possible that simple heuristics during postprocessing could remove these transient errors.
Overall, the POS signals give meaningful predictions for most applications, even in non-face regions.

Figure \ref{fig:result_waveforms} shows predicted waveforms from a 10-second window for the same subject at different locations. The nearest contact PPG waveforms are bandpass filtered and plotted against the predictions to show that many of the predictions contain the underlying dominant pulse even in the presence of higher frequency noise. The difference in waveform morphology across body location shows how informative full-body rPPG can be. For this particular segment, the legs contain a strong second harmonic, which may arise from either the closure of the aortic valve during forward wave propagation, or wave reflections occurring at structural discontinuities along the femoral arteries~\cite{London1999}. Future studies on this dataset will explore different waveform morphologies at a finer scale and their relation to arterial stiffness and blood pressure.

\subsection{Pulse Transit Time}
\label{sec:ptt}

\begin{figure*}
    \centering
    \captionsetup[subfloat]{labelfont=footnotesize,textfont=footnotesize}
    \subfloat[]{\includegraphics[width=0.37\linewidth,valign=t]{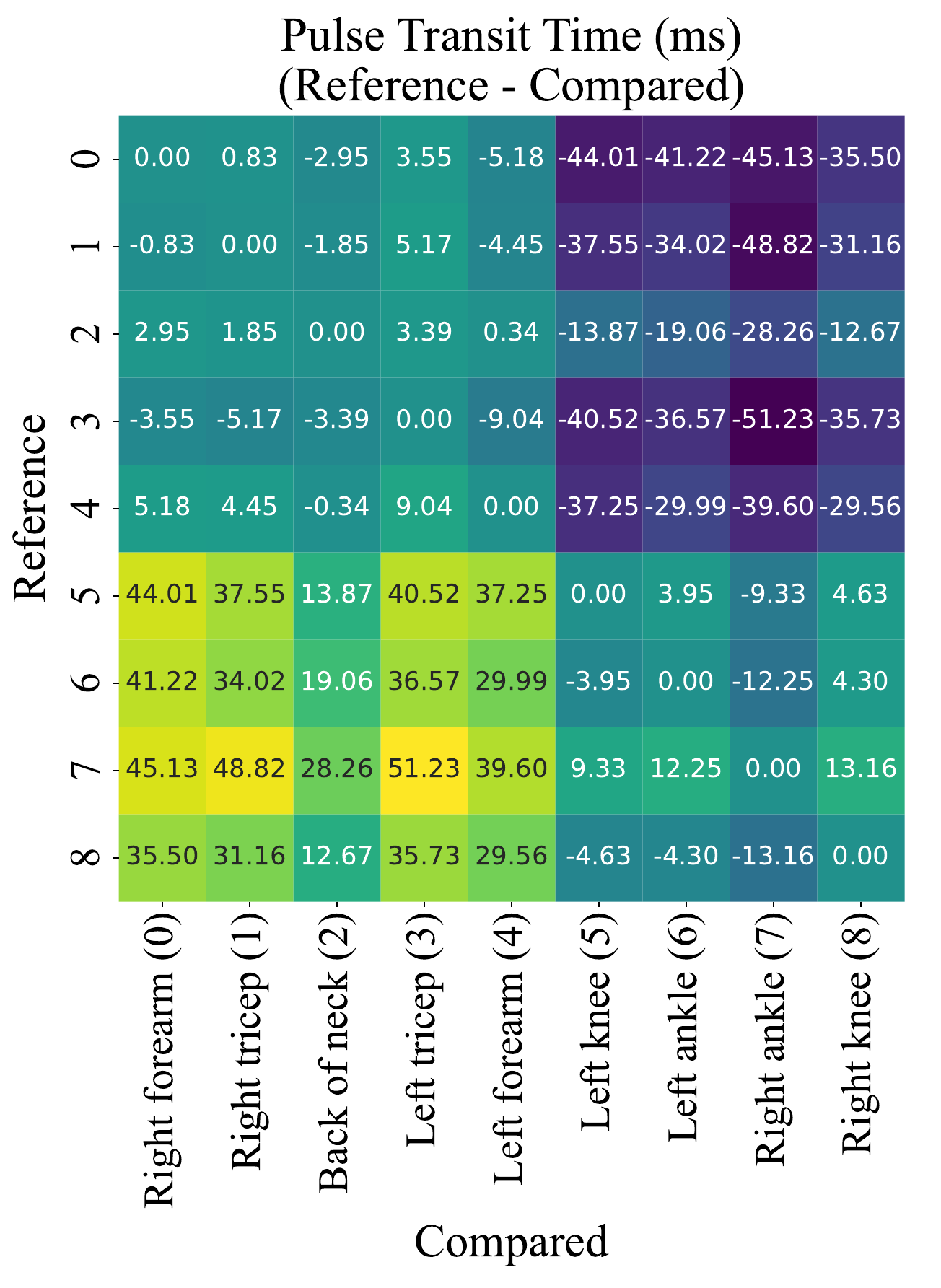}\label{subfig:ptts_}}%
    \hfil
    \subfloat[]{\includegraphics[width=0.25\linewidth,valign=t]{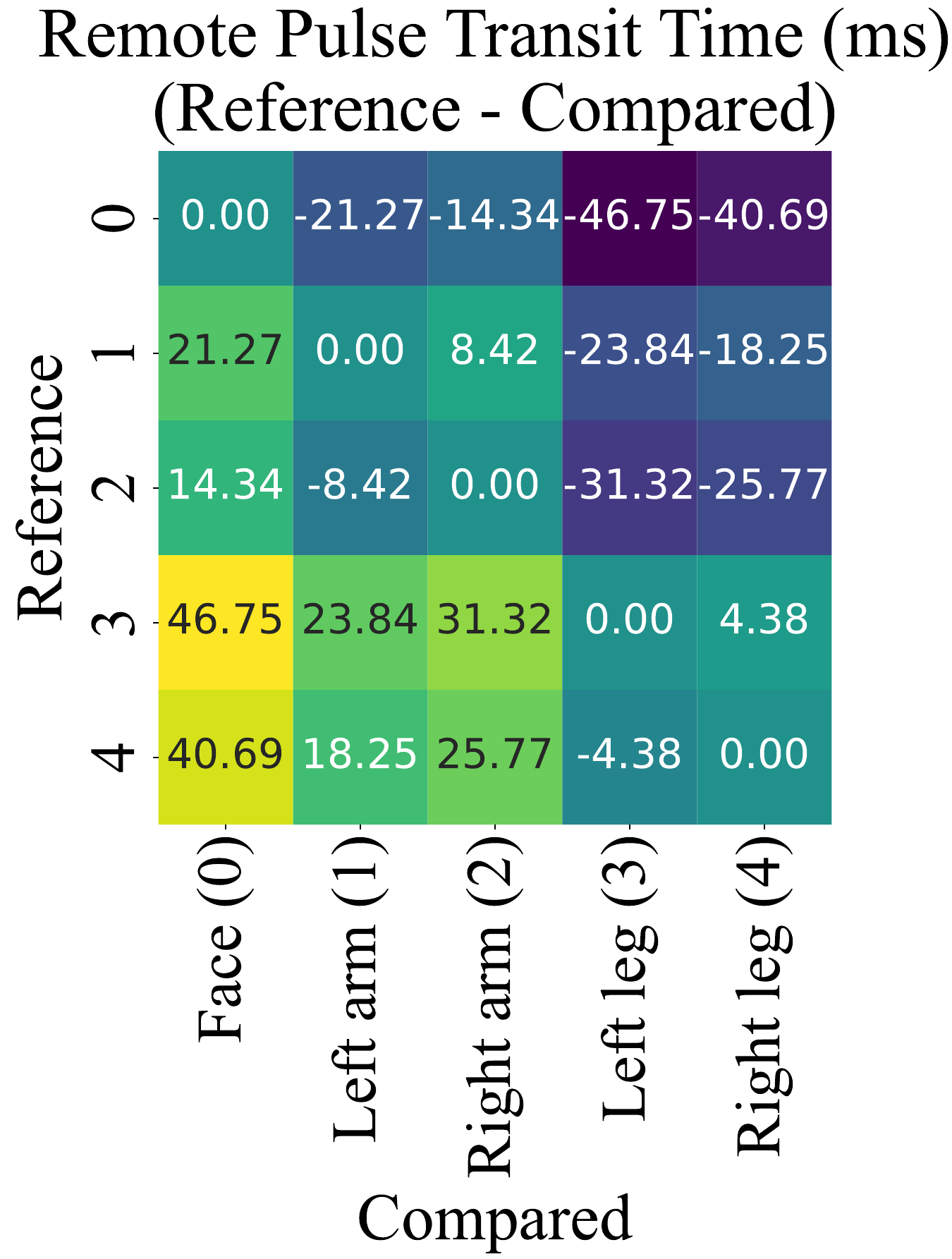}\label{subfig:rptts_}\vphantom{\includegraphics[width=0.37\linewidth,valign=t]{figures/ptt/PTT_subfig.pdf}}}%
    \hfil
    \subfloat[]{\includegraphics[width=0.37\linewidth,valign=t]{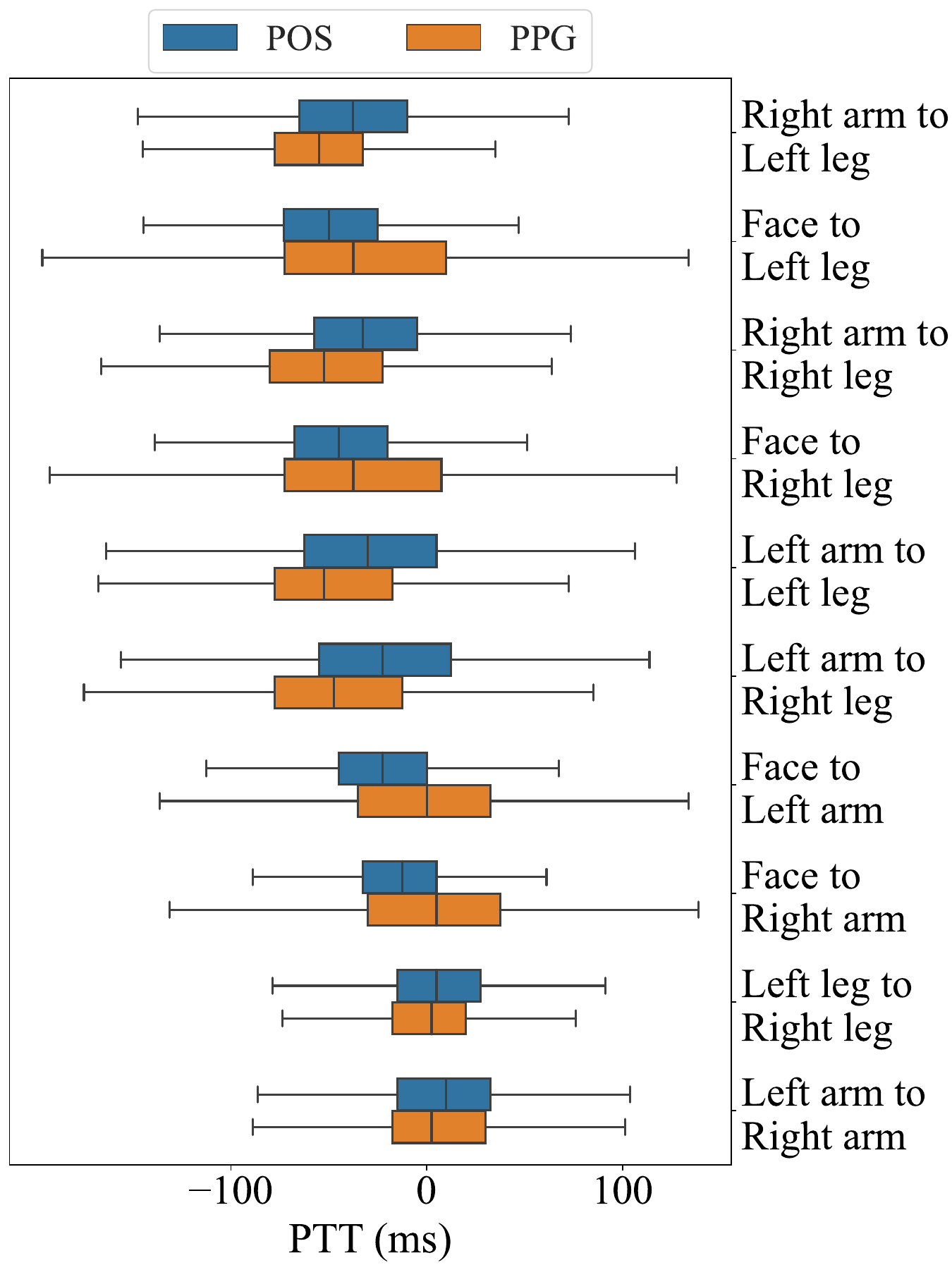}\label{subfig:ptt_boxplots}\vphantom{\includegraphics[width=0.37\linewidth,valign=t]{figures/ptt/PTT_subfig.pdf}}}%
    \caption{Pulse transit time from both contact PPG and rPPG calculated by a sliding cross-correlation between pairs of waveforms. (a) Time lags between the MAX30101 contact PPG sensors. (b) Time lags between POS waveform predictions on the five different body regions. (c) Grouped boxplots comparing the pulse transit time from PPG and rPPG. For the PPG measurements on the arms and legs we only used the forearm and knee sensors, respectively. The PTT is calculated as $X-Y$ given a label of measurement sites ``$X$ to $Y$".}
    \label{fig:ptts}
\end{figure*}

The previous section confirms that non-face rPPG is feasible, which opens new avenues for measuring the blood flow dynamics at multiple locations on the body simultaneously.
We used these results to explore pulse transit time (PTT) from both contact PPG and rPPG signals with results shown in Fig.~\ref{fig:ptts}. A sliding cross-correlation was applied to calculate the phase differences between pulse waves from different sites. We used a window size of 5 seconds (2,000 points) with a stride of 10 milliseconds (4 points), and a maximum lag of 300 milliseconds, which is much higher than a typical transit time~\cite{Block2020}.

\begin{figure*}
    \centering
    \captionsetup[subfloat]{labelfont=footnotesize,textfont=footnotesize}
    \subfloat[]{\includegraphics[width=0.48\linewidth,valign=t]{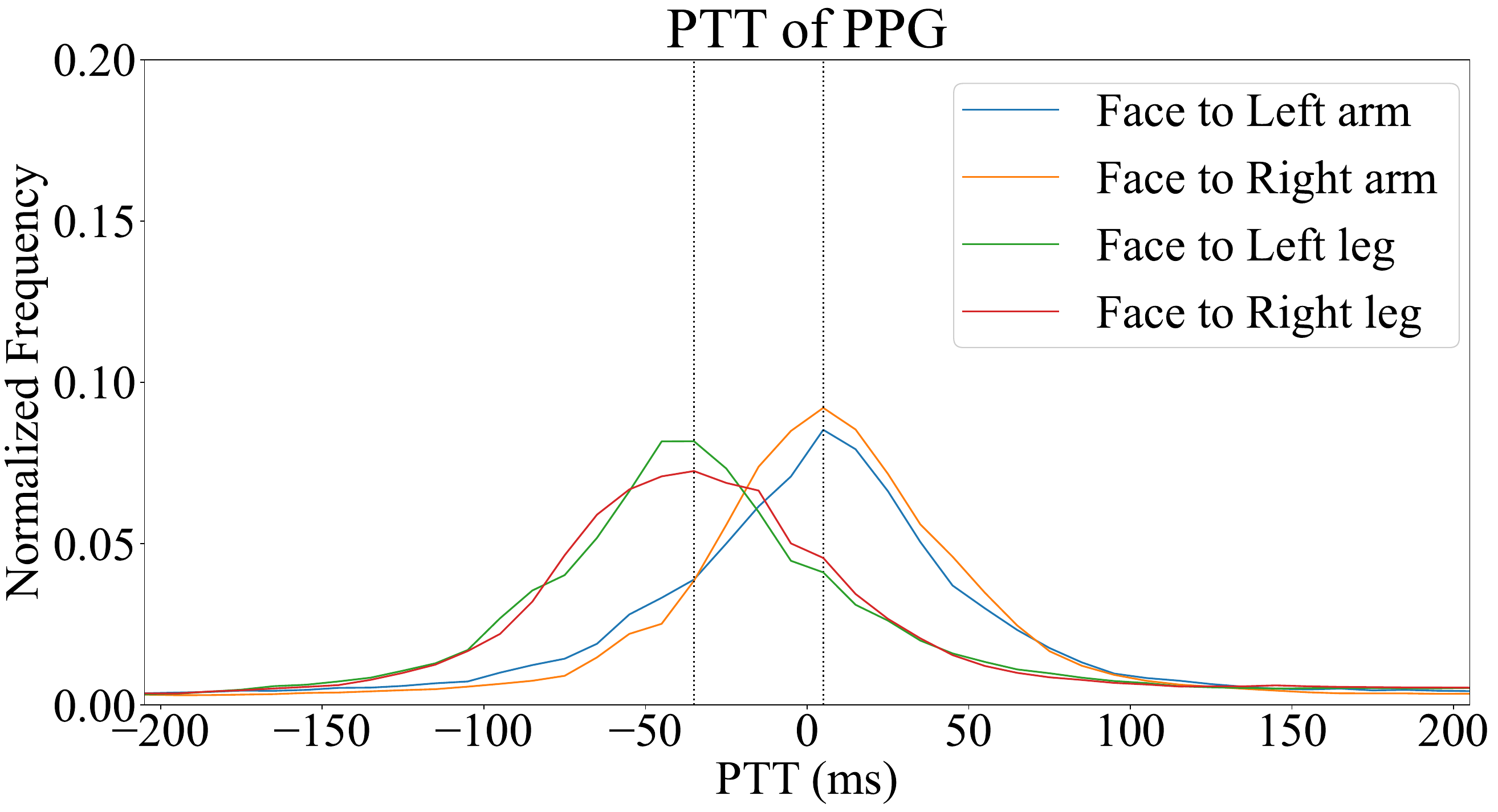}\label{subfig:ptts_all}}
    \hfil
    \subfloat[]{\includegraphics[width=0.48\linewidth,valign=t]{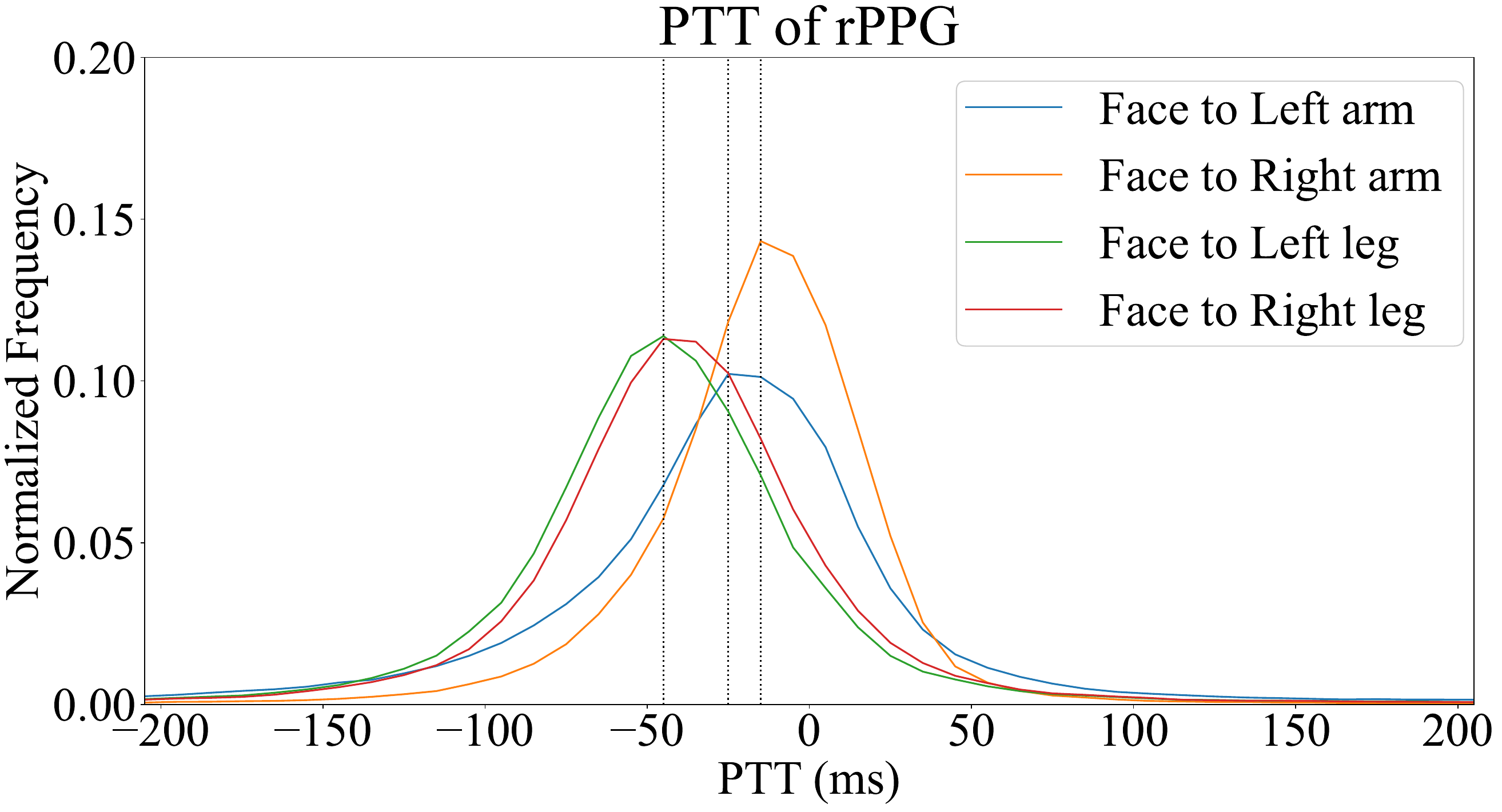}\label{subfig:rptts_all}}
    \hfil    
    \caption{Distributions of PTT and remote PTT among face to left arm, face to right arm, face to left leg, and face to right leg. (a) PTT comparison of all the sites. (b) remote PTT comparison of all the sites. }
    \label{fig:ptts_all}
\end{figure*}

To determine the correlation between the PPG PTT estimates and the rPPG PTT estimates, we chose PTT from face to four sites (both arms and both legs) to explore the comparisons among all the sites, between bilateral sites, and between arms and legs from the same side.
We ignored the component of MSPM containing the adversarial attack, as the pulsating color pattern completely obscured any pulse-related color changes on the face, causing all rPPG methods to fail on that recording segment.
In our analysis, we include PTT estimates up to 200 milliseconds --- which is still much higher than typical pulse transit times at rest~\cite{Shao_2014,Mukkamala2015,Block2020}.
Figure~\ref{fig:ptts_all} shows that PTT and rPTT share similar trends across all of the sites, with PTT from face to arms being generally smaller than PTT from face to legs.

We further observe that the PTT of face to legs for PPG in Fig.~\ref{subfig:ptts_all} is similar to that of rPPG in Fig.~\ref{subfig:rptts_all}, whereas the face to arm distributions disagree.
This could be caused by the sensor placement during our data collection as indicated in Fig.~\ref{fig:sensorplacement}, in which we used sensors 3, 7, 8 and 11 to extract PPG signals. While the leg skin pixels used to extract rPPG signals were mainly from thighs, skin pixels on arms are from both forearms and upper arms, which may account for the systematic offset in this measurement. We additionally observe strong bilateral symmetry in all cases other than face to arms in rPPG data, as shown in Fig.~\ref{subfig:rptts_all}. In this case, the increased variability in left arm rPPG PTT measurements may be due to the blood pressure cuff occluding the upper arms, thus reducing the area of skin pixels used in the rPPG estimation.


\begin{figure}
    \centering
    \includegraphics[width=0.85\linewidth]{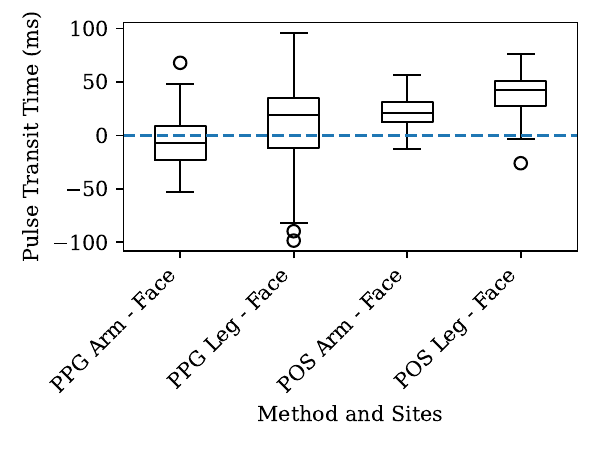}
    \caption{Average pulse transit time differences for each subject between face and arms and between face and legs for PPG and POS methods}
    \label{fig:ptt-test-significance}
\end{figure}

We performed statistical tests to determine the significance of our results, as visualized in Fig.~\ref{fig:ptt-test-significance}. In particular, we first test that the pulse transit time methods using PPG and POS find a greater pulse transit time difference between face and legs than between face and arms, \ie that the Leg - Face mean is greater than the Arm - Face mean for PPG and for POS. Second we test that the PPG and POS methods measure the same phenomenon, \ie that their residuals normally distributed.

\subsubsection{Test of Significance}

We test that the difference in pulse transit time between face and arms and between face and legs is statistically significant. First we perform a Shapiro-Wilk test of normality. For PTT between the face and arms for PPG we calculated a statistic of $0.989$ and a $p$ of $0.702$; for POS the statistic was $0.991$ and the $p$ was $0.0792$. Between the face and legs for PPG and POS respectively the statistic was $0.967$ and $0.974$ and the $p$ value was $0.0258$ and $0.0792$. Since none of these results are significant for Bonferroni corrected $p<0.0125$, we do not find evidence to reject the null hypothesis of normality. Therefore, we proceed to test that results have equal variance using Bartlett's test. We find that neither method yields equal variance in PTT lags between the sites, with PPG yielding a Bartlett statistic of 17.8 (p=$2.47\times 10^{-5}$), and POS yielding a Bartlett statistic of 7.45 (p=$6.33\times 10^{-3}$). Therefore, we use the Kruskall-Wallis test to determine if the means differ significantly.

We obtained the following Kruskal-Wallis results. The average lag difference between arm and leg for each subject differs for PPG with a test statistic of 15.1 (p=$1.00\times 10^{-4}$) and for POS with a test statistic of 42.8 (p=$6.07\times 10^{-11}$). We therefore reject the null hypothesis that the lag difference between arm and leg is equal, rather both PPG and POS methods reproduce the result known from physiology that the pulse transit time difference between face and arm is less than between face and leg.

\subsubsection{Test of PPG/POS Equivalence for PTT}

We further performed a Wilcoxon test in Table~\ref{tab:wilcoxon} on the zero-centered residuals between PPG and POS measurements to determine if they measure the same phenomenon, \ie the residuals are normally distributed. We found no evidence that distributions of residuals from any investigated pairs of sites differ from a normal distribution, suggesting that contact PPG and remote methods like POS are both measuring the same phenomenon when used to measure pulse transit time.

\begin{table}
    \centering
    \caption{Wilcoxon results on residuals between PPG and POS. No results are significant at Bonferroni corrected $p<0.005$}
    \begin{tabular}{c|cc}
        \toprule
        Sites & Statistic & p \\
        \midrule
        Right arm - Left leg & 1847 & 0.777 \\
        Face - Left leg & 1643 & 0.251 \\
        Right arm - Right leg & 1702 & 0.370 \\
        Face - Right leg & 1665 & 0.292 \\
        Left arm - Left leg & 1734 & 0.446 \\
        Left arm - Right leg & 1847 & 0.777 \\
        Face - Left arm & 1869 & 0.849 \\
        Face - Right arm & 1808 & 0.654 \\
        Left leg - Right leg & 1833 & 0.732 \\
        Left arm - Right arm & 1809 & 0.657 \\
        \bottomrule
    \end{tabular}
    \label{tab:wilcoxon}
\end{table}

\subsection{Cross Dataset Analysis}

As many recent rPPG systems employ deep learning, it is important to gauge the value of MSPM as a domain shift. To that end we conduct cross-dataset experiments using datasets that are common in rPPG literature including DDPM~\cite{Speth_IJCB_2021, Vance2022}, PURE~\cite{Stricker2014}, UBFC-rPPG~\cite{Bobbia2019}, and BP4D+~\cite{zhang2016multimodal}. We further compare rPPG techniques including two hand-crafted techniques,
CHROM~\cite{DeHaan2013} and POS~\cite{Wang2017}, along with a deep learning technique, RPNet~\cite{Speth_CVIU_2021}.

\begin{table}
    \centering
    \caption{Intra-dataset results for MSPM, DDPM, PURE, UBFC, and BP4D+.}
    \begin{tabular}{cc|cc}
        \toprule
        Dataset & Method & MAE               & MXCorr            \\
        \midrule
        MSPM & CHROM & 11.012 & 0.134  \\
        MSPM & POS & 9.152  & 0.556  \\
        MSPM & RPNet    & 3.845 $\pm$ 1.518 & 0.704 $\pm$ 0.084 \\
        \midrule
        DDPM & CHROM & 16.223 & 0.076  \\
        DDPM & POS & 9.916  & 0.091  \\
        DDPM & RPNet    & 4.320 $\pm$ 2.877 & 0.553 $\pm$ 0.057 \\
        \midrule
        PURE & CHROM & 2.215  & 0.561  \\
        PURE & POS & 2.323  & 0.565  \\
        PURE & RPNet    & 0.956 $\pm$ 1.097 & 0.741 $\pm$ 0.106 \\
        \midrule
        UBFC-rPPG & CHROM & 3.118  & 0.590  \\
        UBFC-rPPG & POS & 2.671  & 0.598  \\
        UBFC-rPPG & RPNet    & 0.556 $\pm$ 0.219 & 0.808 $\pm$ 0.018 \\
        \midrule
        BP4D+ & CHROM & 3.889 & 0.560 \\
        BP4D+ & POS & 3.290 & 0.565 \\
        BP4D+ & RPNet    & 3.889 $\pm$ 2.227 & 0.566 $\pm$ 0.044 \\
        \bottomrule
    \end{tabular}
    \label{tab:all_intra}
\end{table}

In Table~\ref{tab:all_intra}, we present overall performance for the compared datasets. We used the mediapipe FaceMesh~\cite{kartynnik2019real} tool to extract the signal trace used by the hand-crafted techniques (CHROM and POS). We performed 5-fold cross validation to obtain the mean performance and 95\% confidence interval for RPNet. We observe that the hand-crafted techniques perform similarly on MSPM as on DDPM as opposed to PURE, UBFC-rPPG, and BP4D+. We believe that this is due to similar collection characteristics, specifically with activities designed to elicit a wide range of heart rates. Similarly, RPNet exhibits similar performance across MSPM, DDPM, and BP4D+, with a MAE less than 1 BPM on PURE and UBFC-rPPG, indicating that the three larger datasets may be more challenging than the two smaller datasets.

\begin{table}
    \centering
    \caption{Intra-dataset breakdown of results for MSPM by session portion.}
    \begin{tabular}{c|cc}
        \toprule
        Portion   & MAE                & MXCorr            \\
        \midrule
        full      & 3.845 $\pm$ 1.518  & 0.704 $\pm$ 0.084 \\
        relaxed   & 1.575 $\pm$ 0.453  & 0.729 $\pm$ 0.088 \\
        game      & 1.970 $\pm$ 1.184  & 0.820 $\pm$ 0.047 \\
        breath hold      & 2.228 $\pm$ 0.980  & 0.813 $\pm$ 0.067 \\
        BW video   & 2.588 $\pm$ 3.462  & 0.858 $\pm$ 0.075 \\
        RGB video  & 1.699 $\pm$ 1.096  & 0.834 $\pm$ 0.081 \\
        notattack & 1.820 $\pm$ 0.852  & 0.751 $\pm$ 0.081 \\
        attack    & 14.270 $\pm$ 5.958 & 0.594 $\pm$ 0.064 \\
        \bottomrule
    \end{tabular}
    \label{tab:mspm_intra}
\end{table}

\begin{figure}
    \centering
    \includegraphics[width=\linewidth]{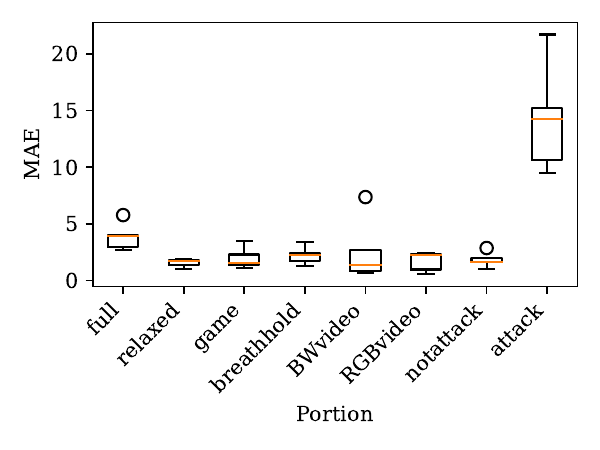}
    \caption{Intra-dataset breakdown of results for MSPM by session portion.}
    \label{fig:breakdown}
\end{figure}

The MSPM dataset contains a sequence of activities separated by relaxation portions as discussed in Section~\ref{sec:collectionprotocol}. In Table~\ref{tab:mspm_intra} and Figure~\ref{fig:breakdown}, we evaluate the performance of RPNet on sub-portions of MSPM. The highest observed errors in the dataset occur during the adversarial attack (during which a red and green pulsating video was displayed on the LCD monitor facing the subject). When this adversarial attack is removed from the analysis, the MAE drops from 3.845 BPM to 1.820 BPM.

We perform a cross-dataset analysis between MSPM and DDPM, PURE, UBFC-rPPG, and BP4D+. In Table~\ref{tab:mspm2all}, we report the performance of models trained on MSPM when evaluated on DDPM, PURE, UBFC-rPPG, and BP4D+. We observe low error rates on PURE and UBFC-rPPG, and relatively higher error rates on BP4D+ and DDPM. For these experiments we trained the MSPM models on the entirety of the MSPM dataset as we found no significant difference in performance between these models and models trained on the subset without the adversarial attack.

\begin{table}
    \centering
    \caption{Cross-dataset results for training on MSPM and evaluating on DDPM, PURE, UBFC, and BP4D+.}
    \begin{tabular}{c|cc}
        \toprule
        Test Dataset & MAE                & MXCorr            \\
        \midrule
        DDPM         & 12.790 $\pm$ 2.949 & 0.361 $\pm$ 0.023 \\
        PURE         & 1.601 $\pm$ 0.664  & 0.667 $\pm$ 0.025 \\
        UBFC-rPPG         & 1.786 $\pm$ 0.339  & 0.690 $\pm$ 0.011 \\
        BP4D+         & 6.111 $\pm$ 0.307  & 0.460 $\pm$ 0.005 \\
        \bottomrule
    \end{tabular}
    \label{tab:mspm2all}
\end{table}

\begin{figure}
    \centering
    \includegraphics[width=\linewidth]{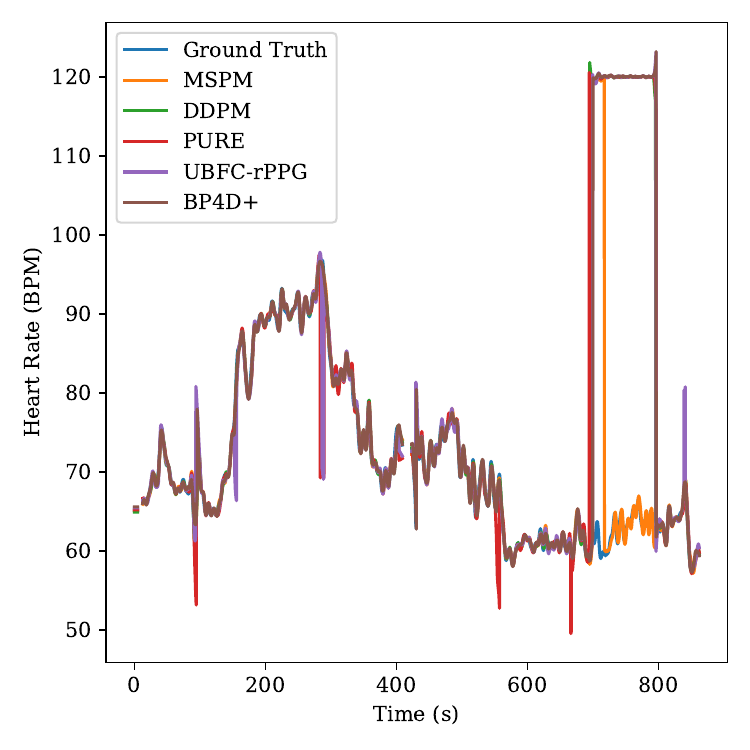}
    \caption{Adversarial attack sample, where the attack frequency of 120 BPM lasts from second 680 to 800.}
    \label{fig:attack}
\end{figure}

In Table~\ref{tab:all2mspm} we report the performance of models trained on DDPM, PURE, UBFC-rPPG, and BP4D+ when evaluated on MSPM with and without the adversarial attack. We find that all cross dataset models obtain a high MAE of over 36 BPM on the adversarial attack, which is over twice the MAE of 14.270 BPM obtained by models trained on MSPM, indicating that training on MSPM yields some level of robustness against this particular attack. An example of an attack is provided in Figure~\ref{fig:attack}, in which a model trained on MSPM (but with this sample withheld) recovers from the attack, while models trained on other datasets succumb. When the adversarial attack is omitted from analysis we observe improved cross dataset performance.

\begin{table}
    \centering
    \caption{Cross-dataset results for training on DDPM, PURE, and UBFC, and evaluating on MSPM.}
    \begin{tabular}{cc|cc}
        \toprule
        Train Dataset & Portion & MAE                & MXCorr            \\
        \midrule
        DDPM & full             & 7.297 $\pm$ 0.133  & 0.647 $\pm$ 0.005 \\
        DDPM & notattack        & 1.705 $\pm$ 0.137  & 0.743 $\pm$ 0.005 \\
        DDPM & attack           & 36.926 $\pm$ 0.066 & 0.150 $\pm$ 0.002 \\
        \midrule
        PURE & full             & 9.269 $\pm$ 0.931  & 0.559 $\pm$ 0.032 \\
        PURE & notattack        & 4.044 $\pm$ 1.142  & 0.632 $\pm$ 0.039 \\
        PURE & attack           & 36.194 $\pm$ 1.280 & 0.199 $\pm$ 0.023 \\
        \midrule
        UBFC-rPPG & full             & 8.283 $\pm$ 0.247  & 0.592 $\pm$ 0.011 \\
        UBFC-rPPG & notattack        & 2.762 $\pm$ 0.251  & 0.680 $\pm$ 0.013 \\
        UBFC-rPPG & attack           & 37.036 $\pm$ 0.372 & 0.159 $\pm$ 0.011 \\
        \midrule
        BP4D+ & full             & 17.087 $\pm$ 6.157 & 0.300 $\pm$ 0.204 \\
        BP4D+ & notattack        & 12.919 $\pm$ 7.096 & 0.357 $\pm$ 0.226 \\
        BP4D+ & attack           & 36.755 $\pm$ 1.003 & 0.114 $\pm$ 0.030 \\
        \bottomrule
    \end{tabular}
    \label{tab:all2mspm}
\end{table}

We find that RPNet trained on MSPM generalizes well to PURE, UBFC-rPPG, and to a lesser extent BP4D+, while it struggles to generalize to DDPM. We also find that models trained on DDPM, PURE, and UBFC-rPPG generalize to MSPM, while models trained on BP4D+ struggle to generalize to MSPM. We believe that this indicates that MSPM provides an adequate challenge as an rPPG dataset.

\section{Discussion}

Several features of the MSPM dataset are under-utilized in our baseline analysis. In particular, we collected but did not report on the NIR video suitable for pupillometry, the utility of the recorded screencast for mitigation of the adversarial attack, and the availability of synchronized video from multiple angles including a profile view. We also did not report results for camera-based blood pressure estimation, which is a potential research application of MSPM. We expect that further research utilizing this dataset will explore these features.
\section{Conclusions}

In this paper we present MSPM: A Multi-Site Physiological Monitoring dataset.
We demonstrated its suitability for multi-site rPPG which enables pulse transit time, a known feature for blood pressure estimation.
MSPM also enables remote respiration, where motion of the chest region in video reveals the breathing rate.
For rPPG, we compared MSPM to popular datasets including PURE, UBFC-rPPG, BP4D+, and DDPM, finding that MSPM provides a challenge for models trained on these datasets, especially through a physical attack in which a computer monitor flashes red and green light from a realistic viewing distance.
We believe that the MSPM dataset will be useful for advancing research utilizing full body rPPG and camera-based physiological measurement.

\section*{Acknowledgment}
This research was sponsored by the Securiport Global Innovation Cell, a division of Securiport LLC. Commercial equipment is identified in this work in order to adequately specify or describe the subject matter. In no case does such identification imply recommendation or endorsement by Securiport LLC, nor does it imply that the equipment identified is necessarily the best available for this purpose. The opinions, findings, and conclusions or recommendations expressed in this publication are those of the authors and do not necessarily reflect the views of our sponsors.

{\small
\bibliographystyle{ieee_fullname}
\bibliography{main}
}

\begin{IEEEbiography}
[{\includegraphics[width=1in,height=1.25in,clip,keepaspectratio]{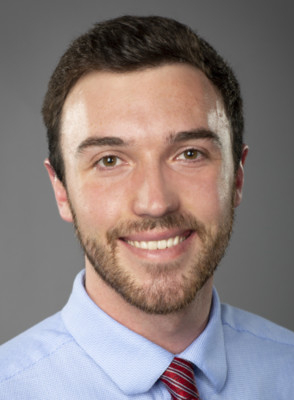}}]{Jeremy Speth} is a fifth year Ph.D. student in Computer Science at the University of Notre Dame, where he received his Masters degree in 2022. Before that, he attained his Bachelor’s degree in Computer Science at the University of Nevada, Reno in 2019. His research interests include physiological measurement, signal and image processing, and deep learning.
\end{IEEEbiography}

\begin{IEEEbiography}
[{\includegraphics[width=1in,height=1.25in,clip,keepaspectratio]{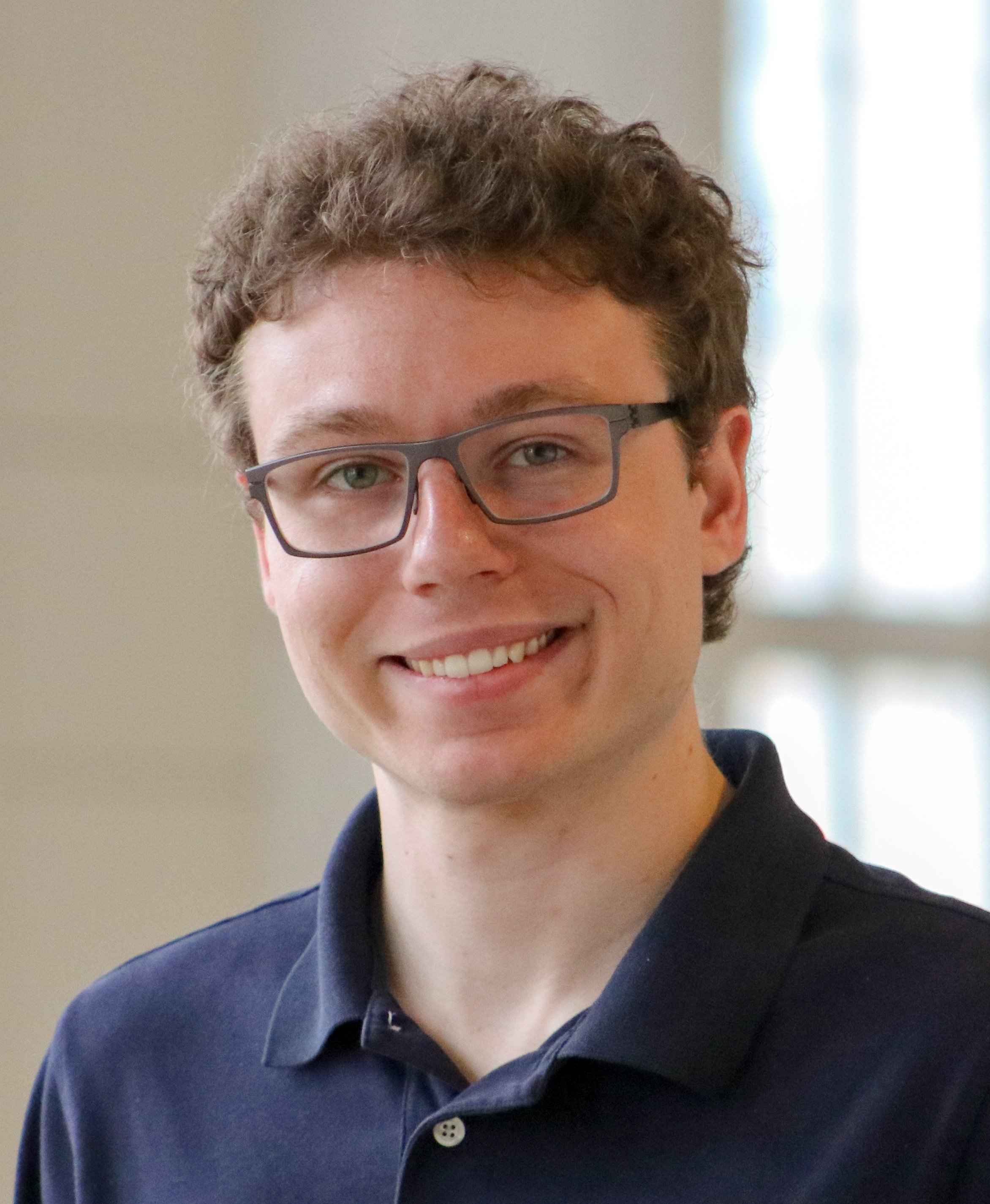}}]{Nathan R. Vance} received a Bachelor of Science in Computer Science and a Bachelor of Arts in Chemistry from Hope College in 2017, and a Master of Science in Computer Science and Engineering from The University of Notre Dame in 2022. He is currently in his seventh year of study in the Computer Science and Engineering doctoral program at The University of Notre Dame. His research interests include biometrics and remote physiological monitoring, including remote photoplethysmography.
\end{IEEEbiography}

\begin{IEEEbiography}
[{\includegraphics[width=1in,height=1.25in,clip,keepaspectratio]{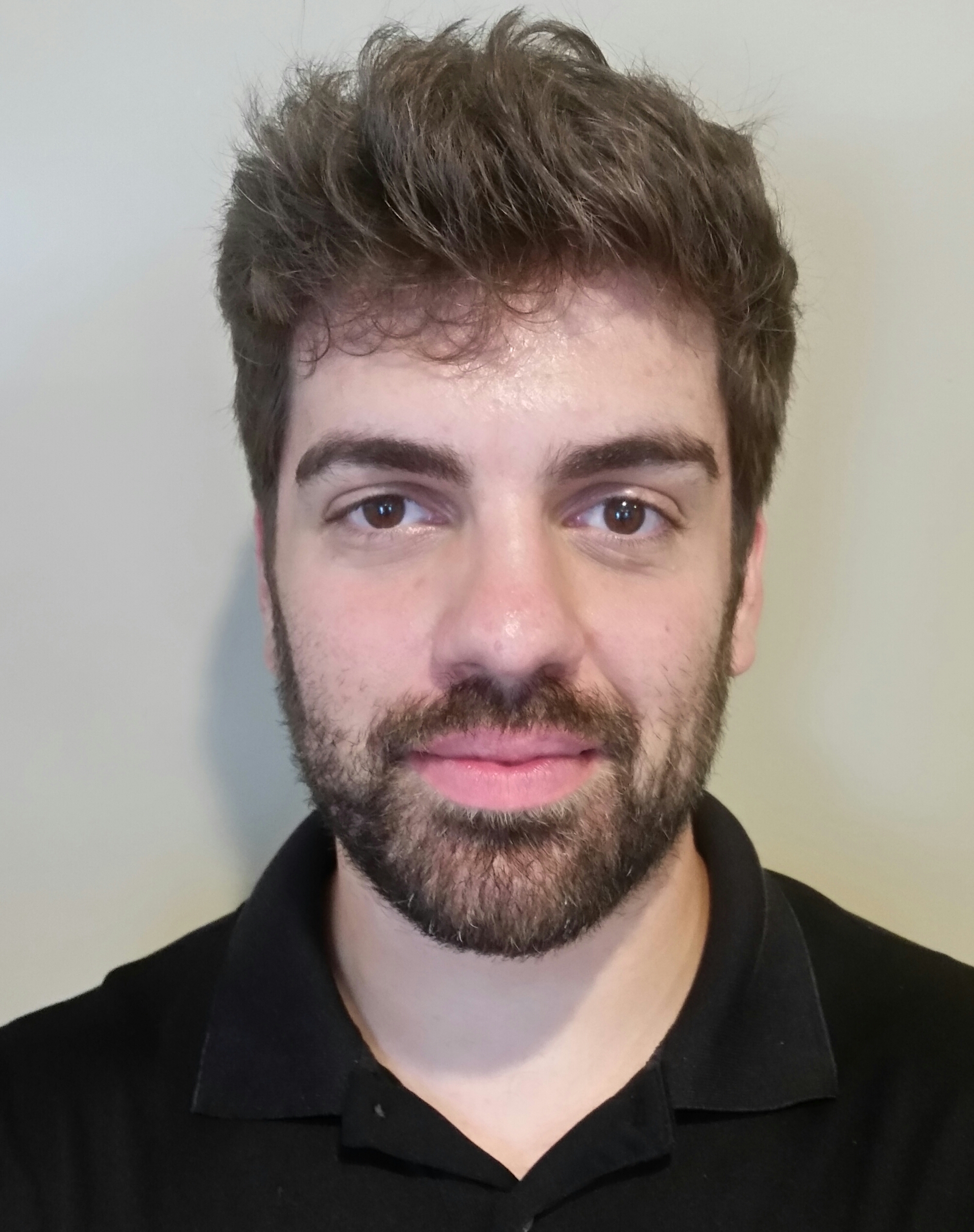}}]{Benjamin Sporrer} is currently in his fourth year of study in the computer science and engineering doctoral program at the University of Notre Dame. He holds a Bachelor's of Science in Mathematics with a co-major in Philosophy from Albright College. His research interests include remote photoplethysmography, trustworthy AI, expert system analysis, and compression techniques. 
\end{IEEEbiography}

\begin{IEEEbiography}
[{\includegraphics[width=1in,height=1.25in,clip,keepaspectratio]{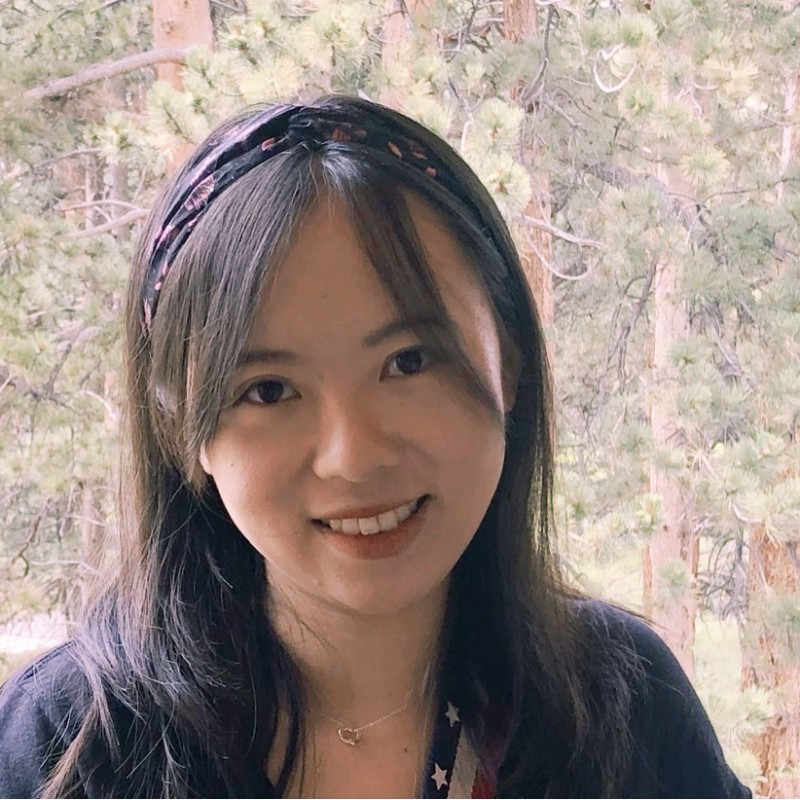}}]{Lu Niu} is a third year Ph.D. student in Computer Science at the University of Notre Dame. She received her Masters of Science in Applied Data Science and Masters of Science in Industrial \& Systems Engineering, both from University of Southern California. She received her bachelor degree in Engineering Management in China. Her research interests include biometrics, image processing, and machine learning. 
\end{IEEEbiography}

\begin{IEEEbiography}
[{\includegraphics[width=1in,height=1.25in,clip,keepaspectratio]{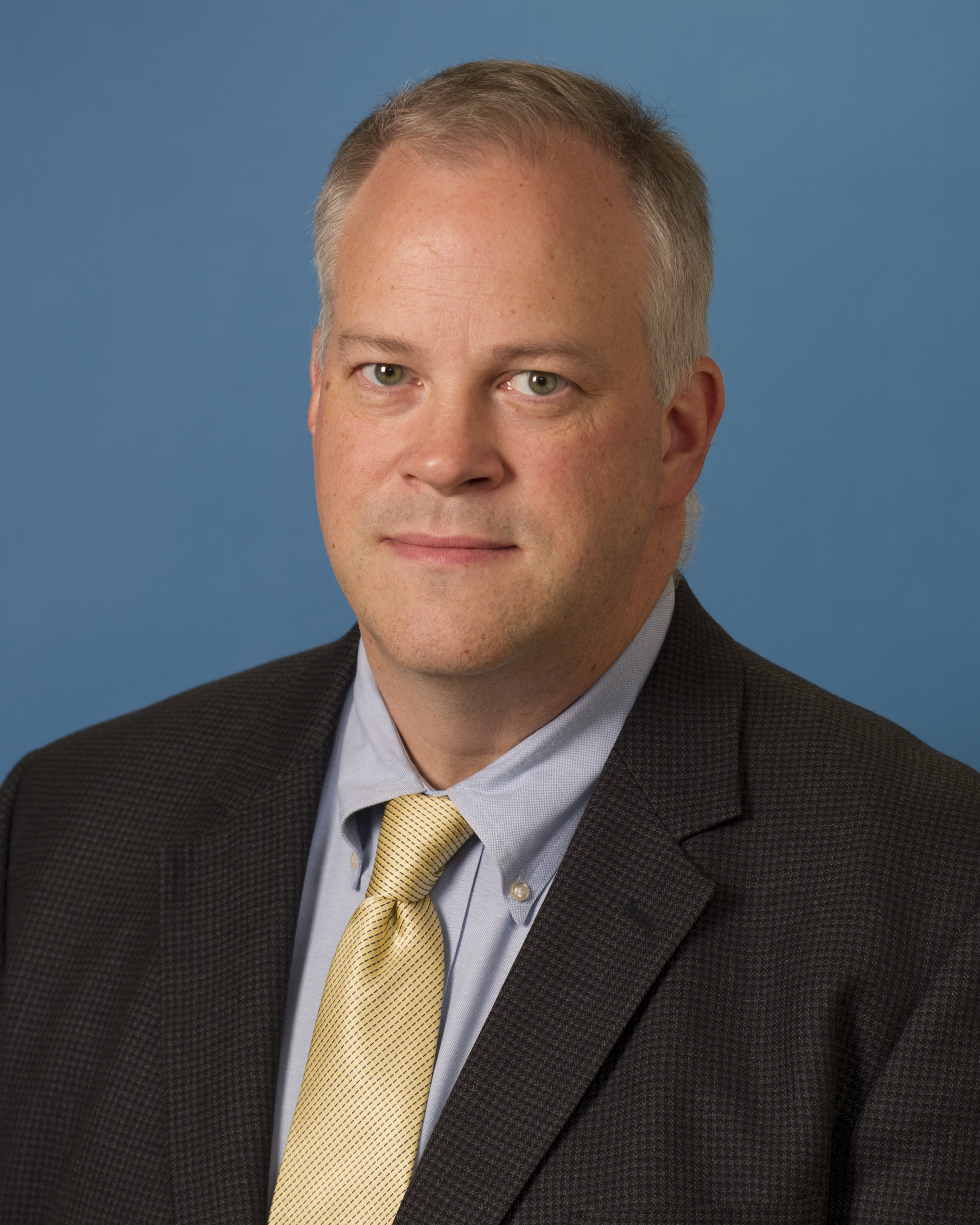}}]{Patrick Flynn} is the Fritz Duda Family Professor of Engineering at the University of Notre Dame.  He received the Ph.D. in Computer Science in 1990 from Michigan State University. He has also held faculty positions at Washington State University and Ohio State University.  His research interests include computer vision, biometrics, and image processing. Dr. Flynn is a past Editor-in-Chief of the {\em IEEE Biometrics Compendium}, a past Associate Editor-in-Chief of {\em IEEE Transactions on PAMI}, and a past Associate Editor of {\em IEEE Transactions on Image Processing} and of {\em IEEE Transactions on Information Forensics and Security}.  He is a Fellow of IEEE and of IAPR. 

\end{IEEEbiography}

\begin{IEEEbiography}
[{\includegraphics[width=1in,height=1.25in,clip,keepaspectratio]{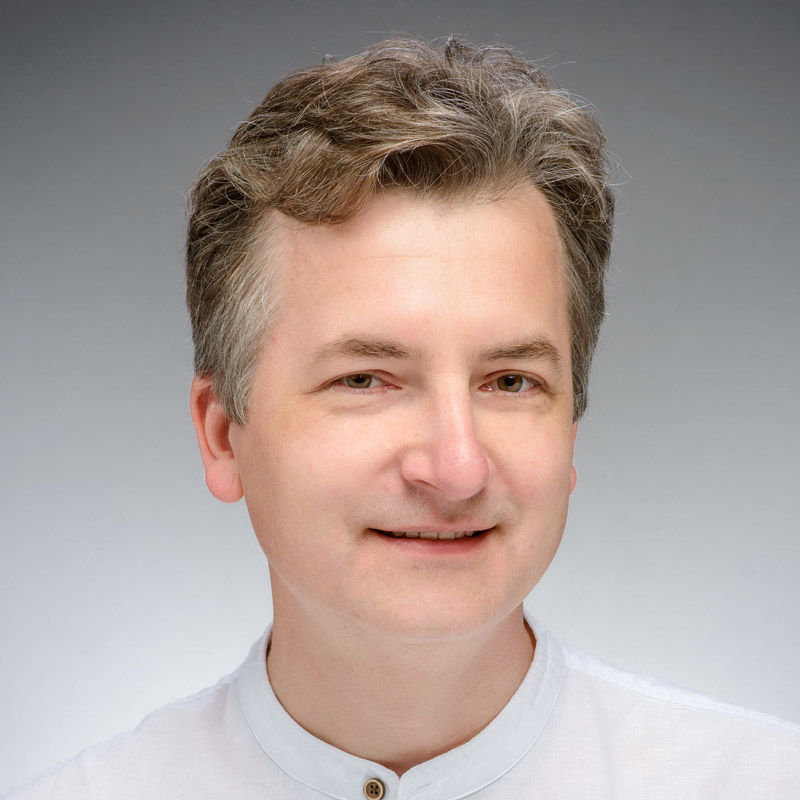}}]{Adam Czajka} is an Associate Professor at the University of Notre Dame and co-directs the Computer Vision Research Lab in the Department of Computer Science and Engineering. His research focuses on human biometrics, especially on iris recognition and methods of detecting biometric presentation attacks. He is the recipient of the NSF CAREER award. Dr Czajka’s research has been funded by the US Department of Defense, US National Institute of Justice, FBI Biometric Center of Excellence, NIST, IARPA, US Army, US National Science Foundation, European Commission, Polish Ministry of Higher Education, and numerous companies.
\end{IEEEbiography}

\end{document}